# Computer Vision and Deep Learning for 4D Augmented Reality.

**Karthik Shivashankar**

*Interim Report for the Master of Science Euromaster in Electronic Engineering*

from the

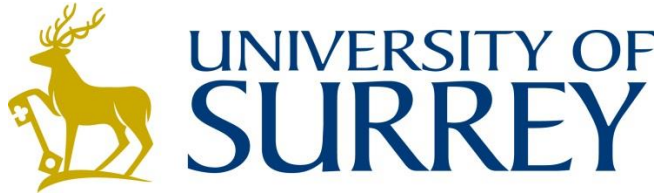

University of Surrey

*Department of Electronic Engineering*

Faculty of Engineering and Physical Sciences

University of Surrey

Guildford, Surrey, GU2 7XH, UK

AUGUST 2019

Supervised by: Prof Adrian Hilton and Dr Marco Volino

©Karthik Shivashankar 2019

# DECLARATION OF ORIGINALITY

I confirm that the project dissertation I am submitting is entirely my own work and that any material used from other sources has been clearly identified and properly acknowledged and referenced. In submitting this final version of my report to the Turn it in plagiarism resource, I confirm that my work does not contravene the university regulations on plagiarism as described in the Student Handbook. In so doing I also acknowledge that I may be held to account for any particular instances of uncited work detected by the Turn it in anti-plagiarism resource, or as may be found by the project examiner or project organiser. I also understand that if an allegation of plagiarism is upheld via an Academic Misconduct Hearing, then I may forfeit any credit for this module or a more severe penalty may be agreed.

Dissertation Title

**Computer Vision and Deep Learning for 4D Augmented Reality.**

Author Name

Karthik Shivashankar

Author Signature       Karthik Shivashankar                    Date: 27/08/2019

Supervisor's name: Prof Adrian Hilton and Dr Marco Volino

# WORD COUNT

Number of Pages:    80

Number of Words:    19148

# Contents

**List of Figures**

# ABSTRACT

The prospect of 4D video in Extended Reality (XR) platform is huge and exciting, it opens a whole new way of human computer interaction and the way we perceive the reality and consume multimedia. In this thesis, we have shown that feasibility of rendering 4D video in Microsoft mixed reality platform. This enables us to port any 3D performance capture from CVSSP into XR product like the HoloLens device with relative ease. However, if the 3D model is too complex and is made up of millions of vertices, the data bandwidth required to port the model is a severe limitation with the current hardware and communication system. Therefore, in this project we have also developed a compact representation of both shape and appearance of the 4d video sequence using deep learning models to effectively learn the compact representation of 4D video sequence and reconstruct it without affecting the shape and appearance of the video sequence.

## 1.0 INTRODUCTION.

Extended reality (XR) is the umbrella term for the augmented reality (AR), Virtual reality (VR) and Mixed Reality (MR). Demand for immersive multimedia experience has propelled the advancement of many XR application and coupled with rapid development in the areas of hardware miniaturization and computing power has enabled users to experience new kinds of realties. These new types of realities include VR, AR and MR. The potential of these technologies will have a huge impact in our day to day life, many tech giants are already heavily investing on their next XR product or service. If we look at the evolution of display technology starting from Cathode Ray tube (CRT) to Thin Film Transistor (TFT) to Light Emitting Diode (LED), just to name a few, it is evident that these technologies are moving towards more immersive and realistic interactive experience. Consequently, the future of media consumption and entertainment will be in a XR platform. However, some of the challenges faced in adapting to XR are cost, usability, VR sickness, poor contents and most importantly the processing power and data bandwidth required for running a reliable complex model in XR platform remains a considerable limitation [1].

4D video technology is key area of active research in Computer vision and graphics community, in a nutshell 4D video technology is a temporally coherent 3D model, which can capture any performance at multiple viewpoints of a 3D object in every frame of reference in a video. This is achieved by capturing the performance of a character using plurality of synchronous camera set up from different view point [2] and feeding the captured data to the 4D video reconstruction pipeline [3]. The potential for 4D video is huge because one can capture the performance of a character in a scene and port them directly into a different reality as a computer-generated model, without having to go through the complex, labour intensive character modelling or animation pipeline. In addition to, it is also possible to render the captured performance or scene in multiple viewpoint for every frame in the video. However, still a lot of research need to be done in 4D video to represent visually realistic and reliable performance capture and to reduce required data bandwidth and overhead that come with the 4D video sequence.

### 1.1 Motivation.

The ulterior motivation behind this project is to test the feasibility of rendering visually realistic 4D video sequence in a XR platform. Few obstacle, in producing reliable visual realistic 4D video sequence are mainly the space and time complexity of rendering a complex temporally coherent 3d models. Generally running a complex temporally coherent model is a computationally intensive task on a mixed reality headset like the HoloLens. In additional to that the mesh and texture sequence occupy lot of space. For some application, they might be a need to stream these 4D video sequence in real time. In such use cases, the data band width and communication overhead required to transmit this information with high fidelity may also pose some serious constraints with the current hardware and communication module.

In this thesis, we aim to eliminate some of these constraints and test its feasibility for rendering 4D video sequence in a Mixed Reality (MR) Headset like the Microsoft HoloLens and to produce compact representation of the both shape and appearance of the 4D video sequence using a Generative Deep Neural Architecture and also being able to reconstruct that compact representation in the Mixed reality (MR) headset like the Microsoft HoloLens. This paves ways to many interesting uses case for 4D video in XR application.

## 1.2 Use Cases

The temporally coherent 3D object in the XR platforms like the Microsoft HoloLens, opens a whole new level of interaction and exciting use case for both XR and 4D video Technology. Some of the use cases are as follows;

- Creating photo realistic animation capable of multi view point rendering, is the next big step in media and entertainment industry. Volumetric capture using multiple camera can render a scene from any perspective that are not bounded to the single camera perspective of the scene.
- Being able to generate visual realistic character animation will be a huge boon for the media industry without the need for Skeleton motion capture and rigging and so on. This technique will expedite the character animation and game play character generation. The implication of rapid character animation generation in XR will enhance the immersive experience and generate a lot of XR content to consume.
- Digital Double is often used in cinema. With the multi view point volumetric performance capture, one can create 3D object of the character. This digital double can be ported into any real and virtual environment  Digital double can be controlled or manipulated in the scene, because it is 3d object. Like any character animation, the photorealistic digital double can be placed anywhere in the scene to perform its role in the movie.
- Immersive 3D surrounding for interaction and training. With HoloLens and 4d video, the level of immersive experiences is multi modal in nature (with sound, Holograms and other spatial mapping, world tag and anchors). Because HoloLens provide various modalities of controls like gaze, voice command and Gesture. The user can manipulate the state of the object, size, rotate and transform and immerse themselves in the digital world. For instance, Scientist in NASA, have created a virtual

Mars for its trainee to explore the Martian surface. It is also used for guided tours, repair or to provide solution for the complex machine or system [4].

- Holoportation, is technique where the digital avatar is ported into the real world, usually for communication between two or more people or avatar. It is 3D capture technology that allows high-quality 3D models of Human motion to be reconstructed, compressed and transmitted anywhere in real time with high fidelity. HoloLens can enhance these capabilities by allowing users to see, hear, and interact with remote participants in 3D as if they are present in the same physical space [5].

### 1.3 Objectives.

In this project, we aim to render a 4D video sequence in a mixed reality headset like the Microsoft HoloLens with the help of Unity Software package and to develop a Deep representation model to create a compact and visually realistic interactive 4D video sequence which is optimized for space and time complexity as shown in Fig 1.

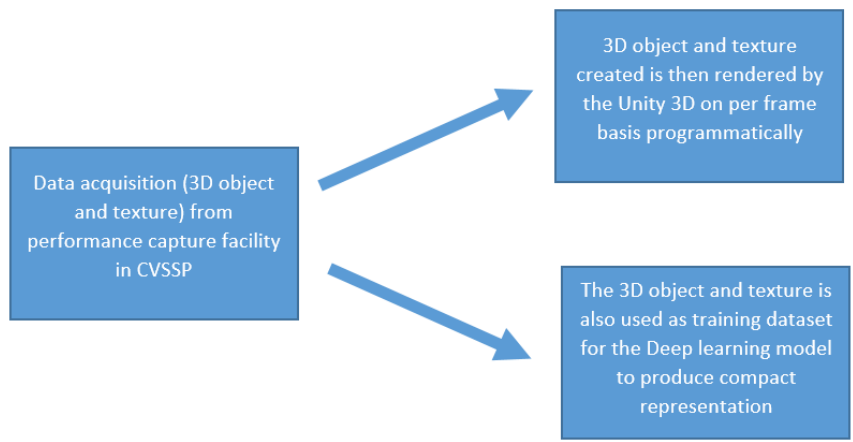

**Fig 1**. Two main work packages for this Project

4D video in Extended Reality (XR) will provide an immersive experience for its users by rendering the temporally coherent 3D models in multiple viewpoint. This is made possible by capturing human performance or any 3D object in a performance capture facility in Surrey CVSSP lab where the 3D mesh and its corresponding texture sequence of the avatar is reconstructed from every viewpoint per frame. The 3D object created per frame is fed to the Unity Game Engine, where the individual frames and its corresponding textures files are applied to create a short clip of the 4D animation programmatically.

Rendering this 4D video sequence using Mixed Reality platform will enable a whole new level of interaction in XR. For instance, this will enable us to capture the human performance and deploying them into the XR platform without having to go through the process of modelling the character and texturing them to creating photo realistic animation, capable for multi view point rendering with high level of user interaction.

3D object created from Surrey CVSSP lab is also used as training dataset for the Deep Learning model to produce a compact representation of the 4D video sequence with reduced Data Bandwidth and optimised for space and time complexity. With the Deep Neural Network architecture we can generate a compact representation of shape and appearance of the 4D video sequence. This will greatly improve the performance constraint for rendering a complex 3D model, by cutting down the latency, redundant data overheads and space occupied by the 3D mesh data structure and texture sequence. This is achieved by using Autoencoder architecture, which we will be discussed in detail in the later section.

1.4 Background on the CVSSP reconstruction Pipeline

Recent advancement in 4D video technology, has enabled us to capture performance of human using plurality of synchronous camera set up [2] as shown in **Fig 2**. CVSSP 4D video construction pipeline uses Model Free (MF) methods, which employs a combination of visual hull and multi view stereo-depth maps segmentation and generating 3D mesh structure for every frame. This generated 3D mesh structure should be temporally consistent over all frame. The temporal coherence for this model free method is proposed by Budd et al [3], this approach uses non-rigid pairwise mesh alignment based on Laplacian deformation mesh framework. Where a collection of unstructured meshes are organised into tree structure based on the shape similarity. The shape similarity tree defines the shortest path of non-rigid surface motion required to align the mesh for every frame over all sequence, creating a temporally coherent 3D model with multiple viewpoints (4D video).

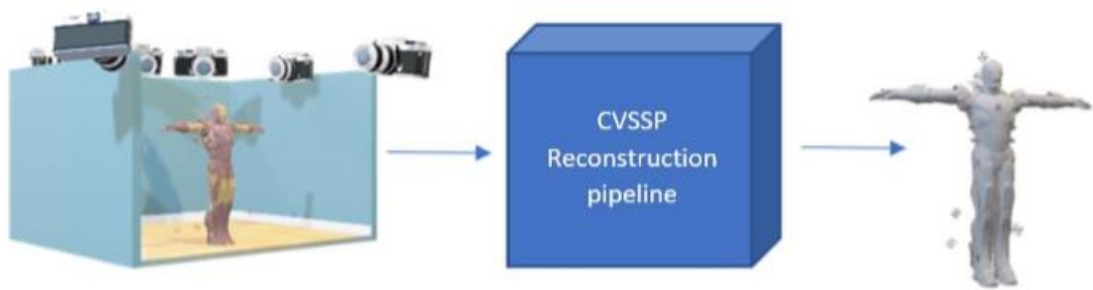

**Fig 2.** Performance capture using CVSSP reconstruction pipeline.

## 1.5 Contribution

Overall contribution made throughout the course of this project

1. Developed a C# algorithm in the Unity platform to programmatically render mesh and texture simultaneously as a 4D video sequence.

2. Deployed the project files into HoloLens device and HoloLens emulator to successfully render 4D video sequence.

3. Created a custom dataset class to train both texture image and its corresponding mesh simultaneously in a mini-batch fashion using Pytorch framework.

4. Implemented a two variants of Variational Autoencoder using a fully connected and Convolution architecture for Mesh 3d co-ordinated and texture sequence respectively and jointly learned the encoding of both the sequence and trained them with various hyper parameter setting and optimisations.

5. Reduced the size of the Mesh and texture sequence from (1024 x 1024 x 3 (Image) + 4000 x 3 (Mesh) = 3157728) 3157728 vector size to 256 latent vector representation, which is roughly $8 \times 10^{-3}$ reduction in size.

6. Generated a decent reconstruction with the trained model from the small training data (around 200 example of training data from each class label). During inference, the model was able to render the Mesh sequence at a speed of 0.00679 per second, which is capable of delivering more than 90Hz refresh rate for a VR or AR application.

## 1.6 Chapter overview

The report is divided into multiple chapters.

Chapter 1

The first technical chapter discusses about the first objective of this project that is to enable 4d video sequence (both Mesh and Texture Sequence) on a mixed reality headset, like the Microsoft HoloLens using Unity's game engine. This chapter is further divided into various sub chapter, wherein each sub chapter will be focussing on achieving some task needed to deliver the final output that is to render the 4D video sequence in the HoloLens platform.

Chapter 2

In chapter 2, we will be discussing about various Deep learning technique to provide the required background for the reader to get the intuition behind the underlying concepts. We will also be discussing about Deep learning on 3D geometry and how is it different from the standard Euclidean data structure(Image).

Chapter 3

In this chapter, we will be primarily discussing about Autoencoder and its variants like the Variational Autoencoder and Conditional Variational Autoencoder. In addition to that , we will discussing about existing Autoencoder architecture citations, that can reconstruct or interpolate 3D geometry and other related works.

Chapter 4

In this chapter, we will explore our second objective that is to develop deep neural network model to generate compact representation of shape and appearance for the 4D video sequence. Here we will be mainly discussing about the datasets used, data pre-processing technique employed and most importantly model architecture, training, experimentation and evaluation of the results.

Chapter 5

Finally, in our last character we will end with conclusion and future directions and references.

## 2.0 CHAPTER 1: 4D VIDEO RENDERING IN HOLOLENS

### 2.1 Background on Unity and Microsoft HoloLens

#### 2.1.2 Unity

Unity software package enable user to control, create and interact with 2D and 3D characters and other multimedia objects. These objects can be assembled into scenes and environments to add lighting, audio, special effects, physics and animation. We can also add interactive character animation and control various logic and deploy them to different platform for production as shown in **Fig 3**.

With Unity game engine, we can also create and render MR reality content at ease using the Mixed Reality Toolkit (MRTK), therefore it is the easiest way to build mixed reality experience for Microsoft HoloLens. MRTK is a special toolkit package made up of many scripts and components intended to accelerate development of MR experience [6].

2.1.3 Microsoft HoloLens Headset:

Microsoft HoloLens is MR headsets capable of augmenting the interaction between the users and computer-generated Hologram or 3D objects and provides a realistic sounds and motion tracking. In addition to that HoloLens offers many interfaces to interact with the 3D virtual objects (Holograms), these objects can interact with the real-world surfaces in our immediate surroundings. We can also attach audio, animation and other components to holograms [7].

2.2 4D video rendering in HoloLens

High level diagram of the System component used in this chapter.

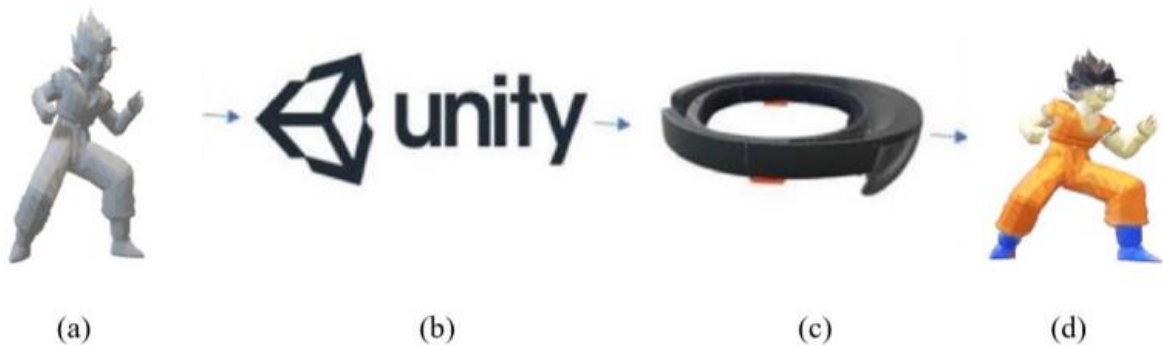

**Fig 3**. Workflow starting from importing, running, exporting and rendering the 3D object into HoloLens using Unity. Left to right (a) 3d object generated from performance capture (b) Importing the 3D object to the Unity; (c) Microsoft HoloLens set up; (d) Render 3d object which is blended in physical environment in a HoloLens Emulator.

### 2.2.1 Importing and running the 3d Mesh using Unity.

Unity comes with a MRTK which makes it easy to render 3D objects in HoloLens. Windows Mixed Reality (WMR) is Universal Windows Application (UWA) and therefore it can directly run on any Windows Platform or system. Below are some requirements and specification need to render 3D mesh and texture sequence in Unity for HoloLens [8].

**Requirements to deploy and render 4D video in HoloLens.**

• Unity with Game development package and Visual Studio 2017 with installed Universal Windows Platform

• HoloLens Device or HoloLens Emulator: To test the 3D virtual object or animation in the local machine before deployment in the actual HoloLens.

**Prerequisite**

• Basic Familiarity with Unity Game Engine

• Visual Studio Editor and C# programming

### 2.2.2 Project Set Up.

To set up a Windows Mixed Reality project in Unity is similar to setting up Unity project for other platforms. However, there are few configurational changes that must be properly set. You need to change the following settings for the project to work on the Windows Mixed Reality[8]. There are as follows.

- Camera Settings
- Script and Resource folder allocation
- Performance Settings

### 2.2.3 Camera Settings

The Camera, that is going to track the Head Mounted Device (HMD) position should be set to the Main camera. This can be done by selecting the camera that we want to track HMD's position and updating it in the inspector view. For HoloLens, the *Clear Flags* property of the **Main Camera** should be set to **Solid**

**Colour** from the inspector view. Also set the **Background colour** to **Black** and the transform position should be set to **(0,0,0)** as shown in **Fig 4**.

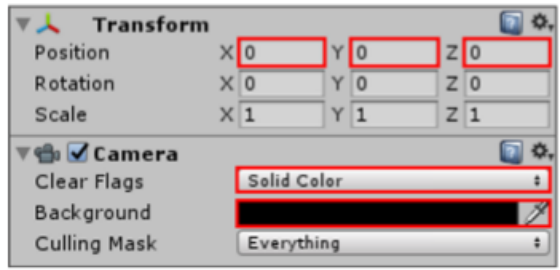

Fig 4. Camera setting for HoloLens compatibility.

2.2.4 Script and Resource folder allocation

All the 3D objects and Texture file assets should be saved the **Resources folder** under the **Assets folder** in the Project view section of the Window as shown in **Fig 5**. Which makes it easier for the C # script to read the assets file in sequential manner and to render the mesh programmatically into the scene. The C# script is used to render the 3D mesh object and texture in a sequential manner programmatically, using the assets located in the Resource folder namely Model and Texture as in **Fig 5**.

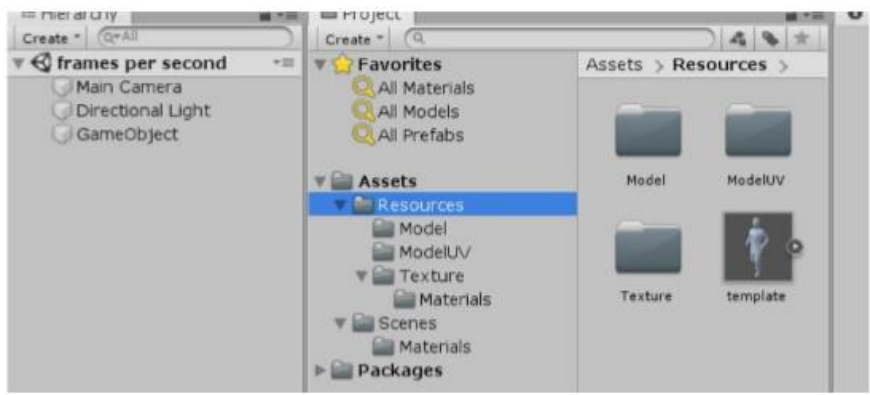

Fig 5. Resource folder allocation and file management

2.2.5 Performance Settings

The HoloLens uses the Fastest quality settings, this can be done under **Edit** > **Project Settings**, then by selecting the **Quality category > Very Low**. The Very Low is the fastest quality setting, which maximizes performance and reduces power consumption for the Headset as shown in **Fig 6**.

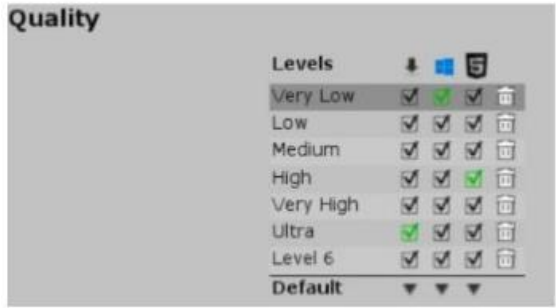

**Fig 6**. Setting the right quality for the Mixed Reality experience.

Another performance optimisation technique that can be used by the Unity is the Rendering setting. Rendering is the process of drawing graphics to the screen or to render texture, there are two stereo rendering for HoloLens, namely

**Multi-pass stereo rendering instance:** Multi-pass rendering runs 2 complete render passes (one for each eye). This generates almost double the CPU workload compared to the single-pass instanced rendering method.

**Single-pass stereo rendering instance:** Instanced rendering performs a single render pass where each draw call is replaced with an instanced draw call. This heavily decreases CPU utilization.

2.3 Algorithm to render the 3D mesh and texture programmatically.

C# is the language that Unity uses to handle all its class and methods. It is the preferred and most widely used scripting language in Unity 3D. In this project, the C# script is used to render the 3D mesh and texture sequence programmatically.

The Pseudocode to create a 4D video in Unity is as follows,

*Input:* Mesh and Texture sequence location, frame_per_second.

*Output:* Programmatically render Mesh and Texture Sequence in the HoloLens Device.

1. Initialise required header and inherit **MonoBehaviour class**
2. Initialise the **Game object and Material object**
3. In the **Start** () method create a **Coroutine** () function
4. **for** L= 1 ……. length (Mesh Sequence) **do**    // Inside the Coroutine function
5. Create **Game object** of type **Mesh render**
6. **Resource. Load (Mesh and Texture sequence location)**
7. Attach texture sequence to the **Mesh render object**
8. **Instantiate the Game object** to render 4D video sequence in the Environment
9. Wait for the **frame_per_second** to complete.
10. Destroy all the Game Object instance.
11. end for

The algorithm initialise the Game object , which is responsible for the scene to be rendered in the environment. And create a co-routine function( )  inside  the start( ) method.  In the co-routine function ,we loop over the length of the mesh sequence , for every iteration in the For loop , the Resource load ( ) function gets the Mesh and its corresponding texture sequence and attaches  these Mesh and Material (texture) instance to the game object , to render them into the scene.Mesh Render and material instance object  are created for ever iteration and waits for specified time before it is all destroyed  for the next iteration to begin and ends when the for loop has reached the length of the Mesh sequence.

### 2.3.1 Publishing settings for HoloLens Device.

The publishing setting is a set of check list with capabilities that you wanted to be included in the HoloLens project set up. Some of the important setting are spatial perception, which provides access to spatial mapping meshes on HoloLens. However, in this project we are not using any of these capabilities, but we have to select **Virtual Reality Supported** under the **XR Settings** and **Enable Depth Buffer sharing** as shown in **Fig 7**.

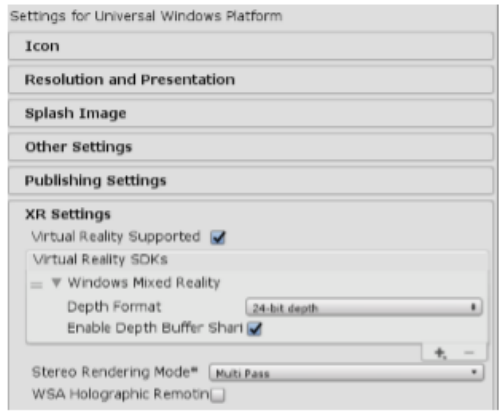

**Fig 7**. Configuring Publishing settings

2.3.2 Building the project for the HoloLens through Visual Studio.

After all the settings are properly configured and when it is ready to be tested, export the visual studio solution. We need to build the Visual Studio solution using Unity for it to be deployed in the HoloLens. Steps to build the solution.

• ***Go to File > Build Settings*** and ***select Universal Windows Platform*** from the Platform list.

• Now click the ***Switch Platform button*** to configure the Editor to build for Windows.

• For HoloLens, you should change the ***Target Device*** setting to ***HoloLens*** as shown in **Fig 8**.

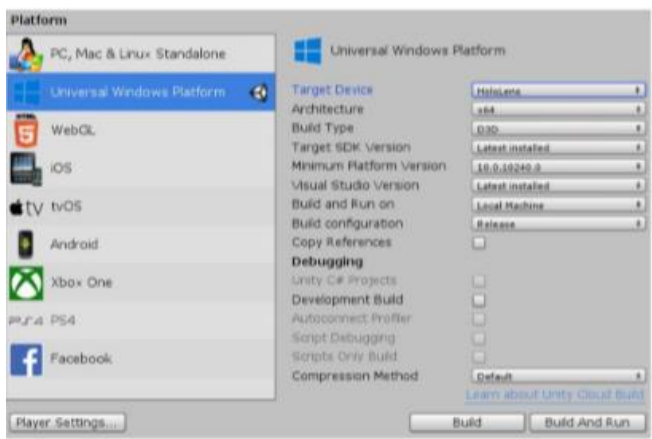

**Fig 8.** Project Build Settings to build the project for HoloLens

### 2.3.3 Deploying the application to HoloLens.

• In Visual Studio, open the *generated solution file* located inside the folder where you *built your project*.

• In Visual Studio menu bar, change the *target platform* for your solution, and select *which device to run the solution on* in this case it is the Device (HoloLens) as shown in **Fig 9**.

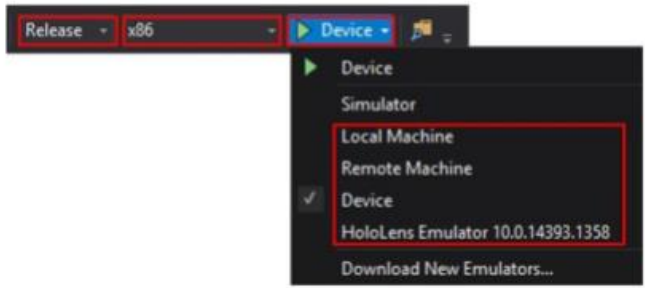

**Fig 9**. Option to deploy the application in HoloLens.

There are three main options to test and render the application in HoloLens from Visual Studio as shown in **Fig 9**:

• Remote Machine • HoloLens Device • HoloLens Emulator.

In this report, we will only discuss on how to deploy the application in the HoloLens Device.

To deploy the application; the *HoloLens Device* should be connected to your *PC via a USB cable*, for the Visual Studio project to be built and deployed into the HoloLens Device.

When we click *Run*, the visual studio builds *the project and install all necessary application* to the connected *HoloLens device* and ready to run automatically once the build process is complete.

### 2.4 4D video output in HoloLens:

After building the project file and deploying it in the HoloLens as discussed in **section 2.3.3**, Later, Upon loading the Unity output file from the HoloLens, you can see a output of a character jogging, this character is captured in the CVSSP performance capture facility. Where the individual meshes and texture of the character are stored in the resource folder, as shown in **Fig 10** below.

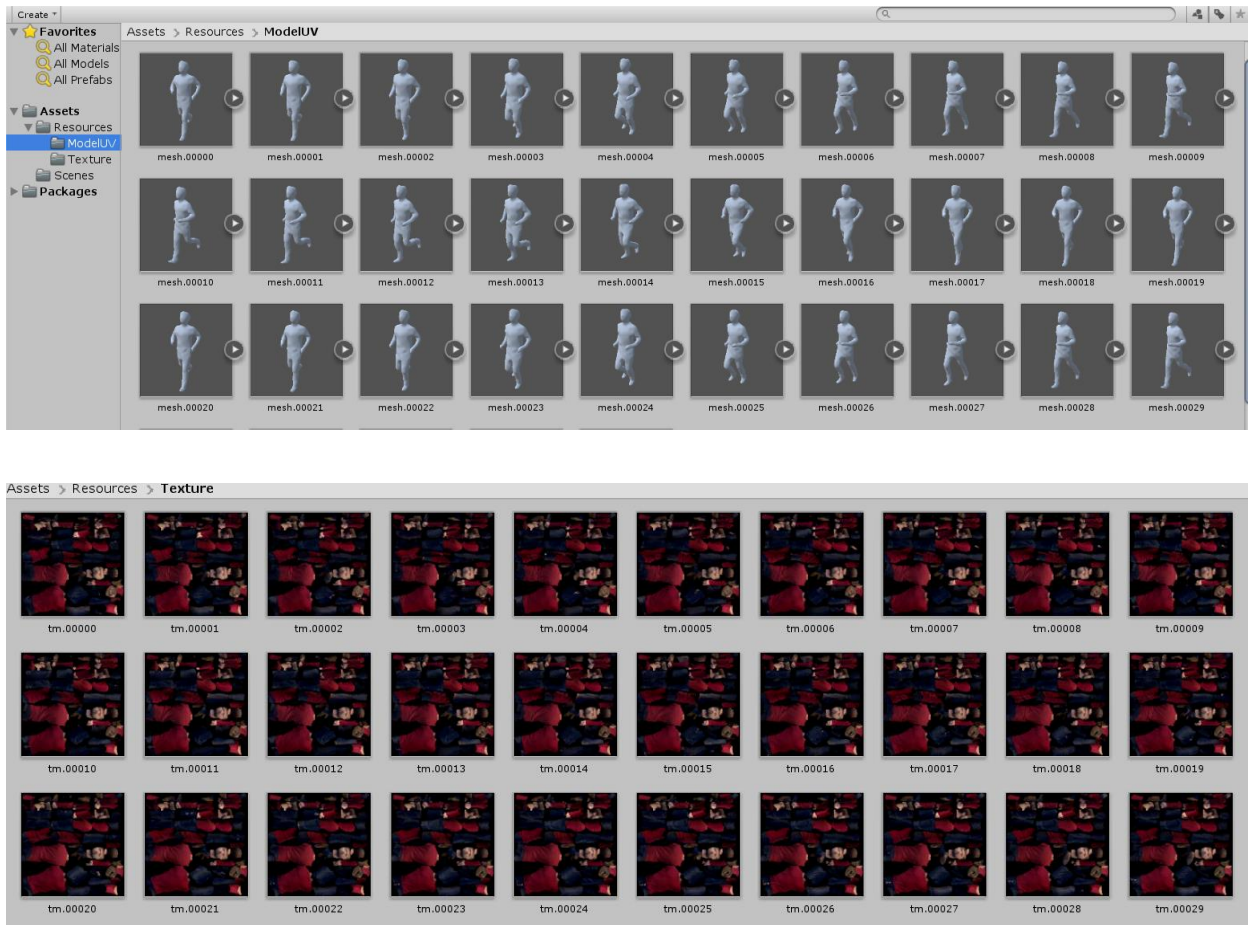

**Fig 10**, The 3D mesh and textures obtained from performance capture stored in the resource folder

The sequence of individual meshes and textures are initialised programmatically using the C# script and renders the character in a frame by frame basis. Which creates the illusion of a video clip or animation. The frame rate can be adjusted to speed up or speed down a character motion. **Fig 11**, shows an individual frame of the character in action. You can also look at the log it generate when a sequence of mesh is rendered programmatically

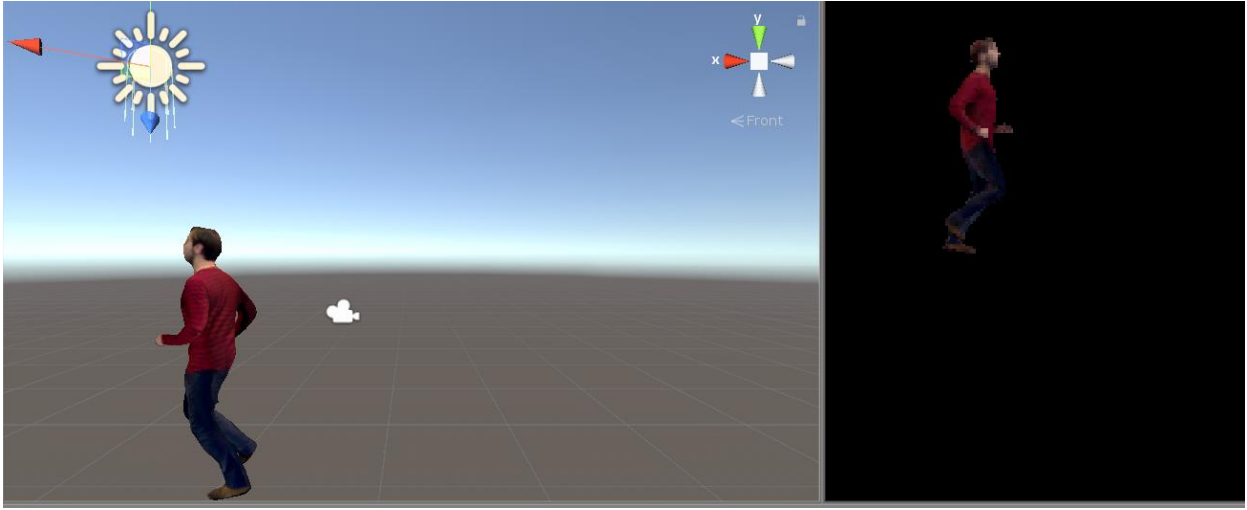

.

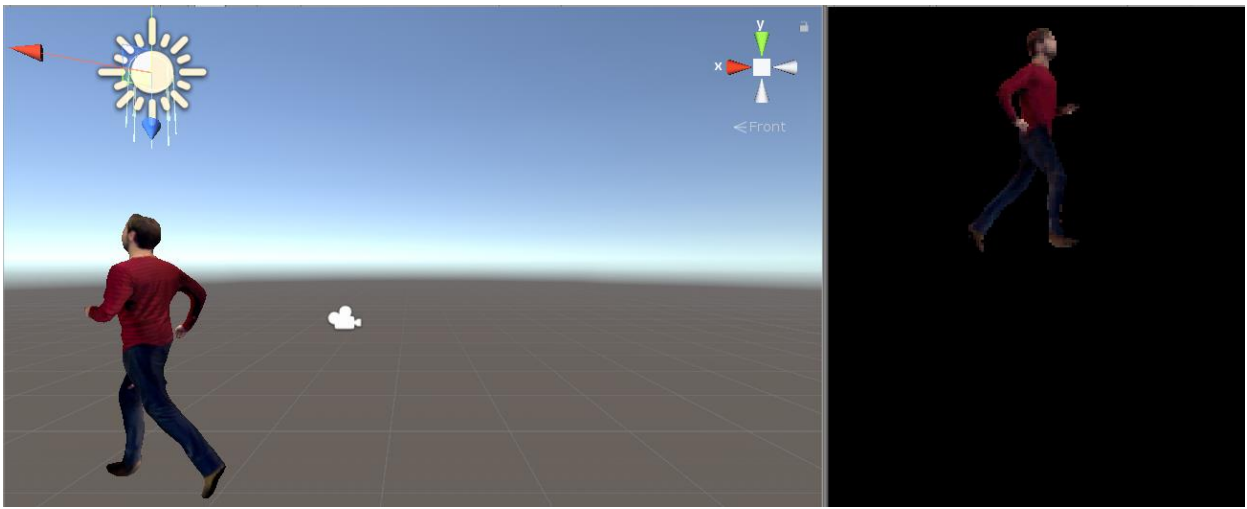

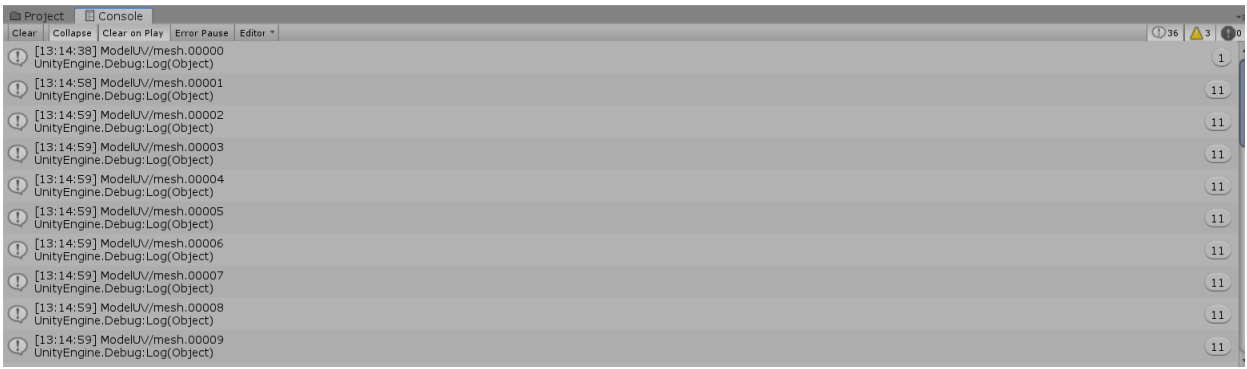

**Fig 11,** Individual frame of the 4D video sequence rendered programmatically and it's Logs.

## 2.5 Conclusion

Rendering the 4D video sequence in the HoloLens is certainly feasible having try it out myself several times during this course of this project and was successful in rendering the 4D video sequence using the C# algorithm I have developed to render the Mesh and texture sequence programmatically. This can be achieved by just setting the right project configuration in Unity 3D platform and deploying the build solution to the HoloLens device.

## 3.0 CHAPTER 2: DEEP LEARNING PRIMER

In this chapter, we will be mainly discussing about Deep learning concepts and develop some intuition about the subject. This will help us understand the role deep learning will play in producing a compact represent of 3D model. We will also discuss about the Deep learning for 3D geometry.

### 3.1 Background on Deep Learning for 3D

Deep learning (DL) is a sub field of Machine learning, which has gained rapid momentum in the last seven years, In a nutshell, deep learning consist of several layers in between the output and input layer, which allows the DL architecture to learn different representation of the input data ( features ) . The non-linear function in the different hierarchical representation is used to learn the underlying pattern and features.

However, most of the DL architecture is designed to work with Euclidean- structure data like image (2D) data. Generally, DL architecture require large quantity of data to effectively learn the underlying feature and generalise to similar set of input features. DL model are generally data driven, over the past few years we saw the drastic rise in low cost 3D acquisition technologies like Microsoft Kinect and Intel Real sense 3D and many more. The tremendous availability of 3D data provides rich data set for building our 3D DL architecture.

3D data has many representations, where they vary in structure and geometric properties from one and another. 3D data are classified into two main categories.

3D Euclidean data has underlying grid structure that enables us to train them using the existing DL architecture like the CNN, RNN and another DL model. These data structures are more suitable for analysing rigid bodies, which are prone to less deformation [9]. Example of 3D Euclidean data structure are Descriptor, Projections, RGB-D, Volumetric (Both Voxel and Octree) and Multi-View images as shown in **Fig 12**.

On the other hand, 3D Non-Euclidean data, do not have any gridded data structure, which make very challenging to train using the tradition DL techniques. Nevertheless, 3D Non-Euclidean are more robust and versatile to non-rigid object and can handle deformation [9]. Which makes it ideal for our use case. Example of Non-3D Euclidean are Point clouds, Graphs and Meshes as shown in **Fig 12**.

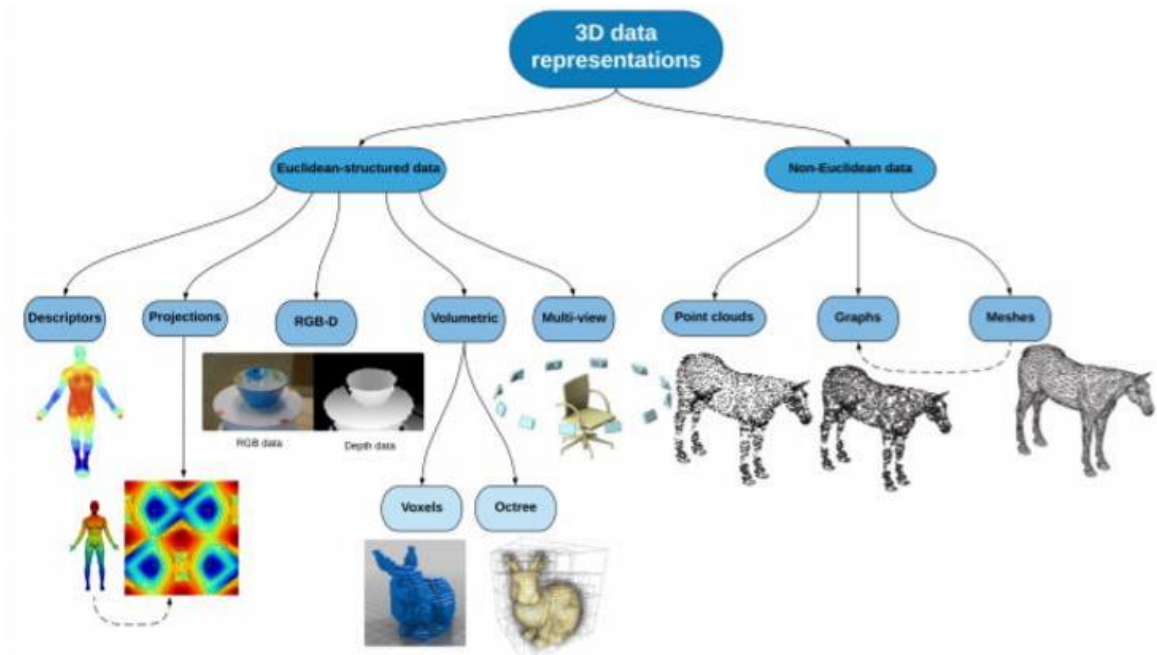

**Fig 12**, Classification of 3D Data Representation [9].

Deep Learning on Euclidean Domain allows the model to learn raw data and to automatically produce representation needed for regression and classification. The model learns multiple level of representation of the input data, obtained by composing non-linear methods at different level in the multilayer network. Deep learning has gained lot of momentum, ever since the Convolutional Neural Network (CNN) won the ImageNet Competition in 2012 [10]. The success of the NN goes to their ability to leverage statistical properties of the data. These properties are effectively utilised by the Convolution Neural Networks (CNNs). CNN extracts the local stationarity of the input data or signal across the data domain. These feature or activation maps are identified by local Convolution filters or kernels. While CNN is powerful tool, which is widely used in different machine learning algorithm, we cannot however use it in non-Euclidean domain like 3D mesh or graph data structure. This makes it challenging to implement NN architecture in 3D model in non-Euclidean Domain.

It is evident from various research that 3D non-Euclidean data structure is best suited for Human action (non-rigid) and can handle large scale deformation [9]. Consequently, we will explore NN architecture in Non-Euclidean data structure like Point clouds, Graphs and Meshes. Geometric Deep learning is the term used for Deep Learning in non-Euclidean domain.

## 3.2 Deep Learning for 3D Shapes.

Deep Learning on 3D shapes has gained immense popularity recently some of the major related work are as follows, Boscaini et al. [11, 12] generalised CNNs from the Euclidean domain to the non-Euclidean domain, which is useful for 3D shape analysis, by establishing correspondences. Brostein et al [13], utilising CNNs on non-Euclidean domain like graphs and meshes, Masci et al.[14] proposed the first convolution operation on mesh, by applying convolution filters on local patches represented in geodesic polar coordinates. Sinha et al.[15] converted 3D shapes to equivalent geometric image to obtain a Euclidean parameterisation, on which CNNs can be implemented. Wang et al.[16][17] Proposed octree-based convolution for 3D shape analysis.

Many related works are also studied convolution operation using vertex features as inputs, The preferred choice of mesh data set is to make the mesh data structure has same connectively but different geometry. Tan et al. [18] use Convolution Autoencoder to extract localised deformation components from mesh datasets. Gao et al. [19] proposed a network which combines convolutional mesh VAE with CycleGAN [20] for automatic unpaired shape deformation transfer. Recently many related works as showed us, that CNNs extended to irregular graphs in spectral domain[21] performed significantly better than the convolution in the spatial domain[18,19]. However, it is important to note that there is no proper pooling operation in non-Euclidean domain [22, 19], consequently the existing mesh based VAE could not aggregate all the local neighbourhood information. Some Recent work has explored this problem and provided a simplification algorithm which directly drops the vertices and uses the barycentre in triangle to recover the lost vertices by interpolation [23]. In order to better represent 3D meshes, a straightforward approach is to use vertex coordinates of a 3D shape. However, vertex coordinates are neither translation invariant nor rotation invariant, making it difficult to learn large-scale deformation. In the next section, we will be going through the basic building block of neural network and its layers and other optimisation function which will help the model to effectively learn the features.

## 3.3 Deep Learning

Deep Learning or Machine learning in general are algorithms that can figure out how to learn program or features or important task by generalizing from the examples (training data). This is attractive alternative to manually constructing the rules for the algorithm to generalize to all variation of a given example.

As more data becomes available, we can tackle more ambition problems through Deep Learning. After all we are in information age, where the value of data overtook the value of oil. Data is abundant and many research institutes have open soured their dataset for the Machine Learning communities to research, learn and to innovate. We can also find the repositories of such datasets, one such example is the UCI Machine Learning Dataset [24].Deep learning or Machine learning in general can learn many different kinds of task, the three main class of algorithm are Classification, Regression and Generative [25].

Three main components for any Machine Learning Algorithm are as follows

**Representation:**

Representation of the input and model play a vital role in deciding what type of task the algorithm is going to learn and how effectively it is able to generalise to the training and test examples. For instance, depending on the task on hand, the representation of the model could be Neural Network Architecture or K-nearest neighbours or Decision trees and so on.

**Evaluation:**

The evaluation is also called objective function, as the name suggest it is used to evaluate the model performance, For instance some of the well know evaluation function of a machine learning algorithm are as follows , Accuracy , precision and recall and Cost function.

**Optimisation:**

Finally, choosing the right optimisation technique to update the learnable parameter like weights and bias in a Neural Network architecture. The most popular and in fact the only optimisation technique for Neural network in general is the Gradient decent algorithm, which is used to find the best fit of learnable parameter like weight and biased depending on the training example and also should be able to generalise to previously unseen data (test example).At the end of the day there is no single recipe in choosing the Representation, Evaluation and Optimisation algorithm.

The ultimate objective of any machine learning algorithm is to generalise beyond the example in the training sets. However, there is one consequence after achieving the generalisation in a Neural Network that we will not have access to the function we want to optimise [25].

**Feature engineering**

Feature engineering is the main ingredient when building the deep learning model, Deep learning model is only good as the data we provide. Learning from the data should be relatively easy if there is a high correlation between the independent variation (features or example) with a given a dependent variable (class or output)

Often time's significant amount of time and effort is spent into feature engineering, rather than actual machine learning. Feature engineering processing includes gathering relevant data, integrating it, clean it and pre-processing it. It is not a one-shot process, it has to be iteratively optimised or modified based on the evaluation result. Domain specific knowledge also pivotal while engineering the feature based on the Deep learning task at hand [25].

### 3.3.1 Neural Networks

Neural networks are collections of algorithms, which mimics the functioning of human brain. Like human brain, it trains to recognise pattern from huge volume of training data and then uses the knowledge to generalise or deduct to another similar scenario.

The architecture of typical NN model is made up of multitude of interconnected simple processing unit called perceptron or artificial neurons as shown in **Fig 13**. NN is usually composed of one or more hidden layer of perceptron called Multi-Layer Perceptron (MLP), which usually consist of an input layer, one or more hidden layer and an output layer. Each hidden layer is made up of number of perceptron which are fully connected to consecutive layer with a weighted parameter, which gets updated for during training [25, 26].

### 3.3.2 Building block of Neural Networks

**Perceptron**:

Perceptron are the basic building block of the neural network, it was biologically inspired from firing action of the Human brain neuron, was developed by scientist Frank Rosenblatt.

**Mechanics of perceptron:**

In a nutshell, a simple perceptron acts like a filter, which is selectively permeable to select range of inputs depending on the threshold value. It takes many binary inputs and yield a single binary output. The below representation shows the toy example off simple perceptron with inputs x1, x2….and yields a single binary output.

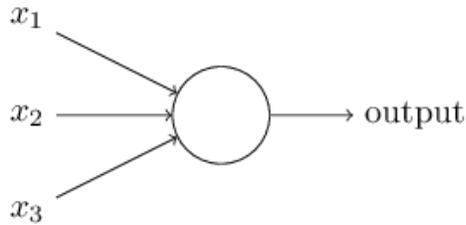

**Fig 13,** Simple perceptron representation.

Rosenblatt introduced the idea of applying weights to perceptron, which expresses the significance of a certain input features. The output 0 or 1 produced by the neuron is determined by the weighted sum $\Sigma \omega_j x_j$ is less than or greater than threshold value set as shown in equation 1.0.

$$output = \begin{cases} 0 & if \sum_j WjXj \leq threshold \\ 1 & if \sum_j WjXj > threshold \end{cases} \quad (1.0)$$

By varying this weight and the threshold, we can construct different logic gates to build a complex mathematical model [26].

Multi linear perceptron.

Multi Linear perceptron is commonly known as MLP, each layer in the MLP is stacked by perceptron. MLP is made up of multiple of these linearly stacked perceptions called layers (generally known as Hidden layer) as shown in **Fig 14**. Multiple Perceptron can be stacked into layer to build a complex model, as the number of layers in the perceptron increases the complexity, learnable parameter and decision-making ability also increased [26].

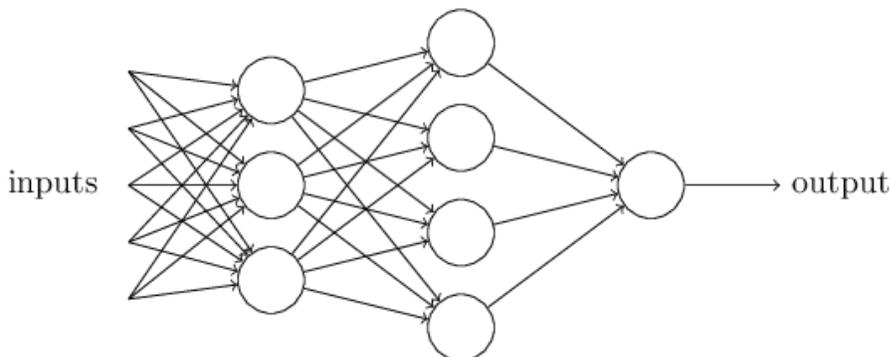

**Fig 14**. Structure of Multi-Layer perceptron

## 3.4 Convolutional Neural Networks

Convolution Neural Network (CNN) is a Deep Learning algorithm which learn to capture the spatial and temporal dependencies from the input data. CNN are best explained with images, the architecture of the CNN shown in **Fig 15**, is similar to that of the connectivity pattern of Neuron in the Human brain and was inspired by the visual cortex. A certain collection of neurons in the visual cortex responds to the certain pattern from the visual field called the receptive field. Similarly, while training, CNN model can take an input image and assign weights and bias to various receptive field by convolving the image with a kernel (also called as filter). These kernels are learnable filters and are generally small in dimension [26].

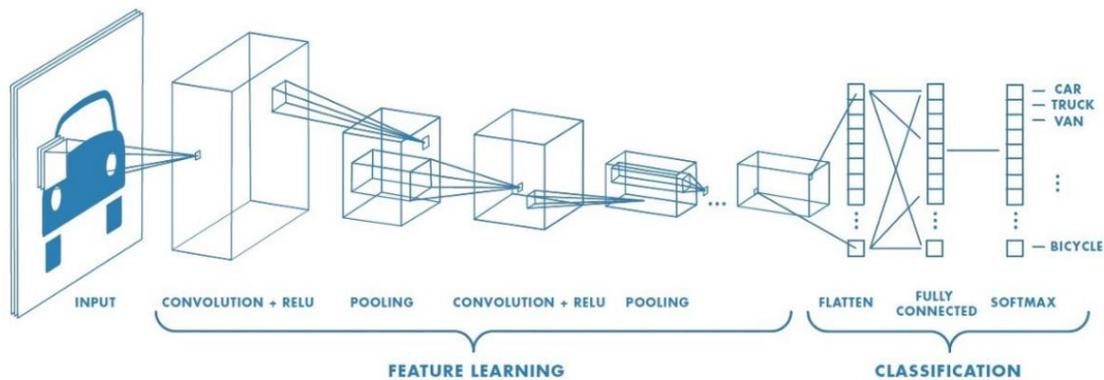

**Fig 15.** Typical CNN Architecture.

Convolution layer is nothing but element wise multiplication of the local region in the input data with kernel weights or model parameter along every depth dimension. Depth of the convolution layer corresponds to the number of kernel filter to use, to learn different activation map in the input by convolving from the top left corner of the image receptive field to the bottom right of the image. The number of steps it takes to slide over the input image is decided by a hyper parameter called stride. For instance, a stride of 2 will reduce the inputs (height x width) by half, if no zero padding is applied. Zero padding, as the name suggests, it pads the input volume with zeros around the boundaries, which also helps in controlling the spatial dimension of the output volume.

The choice of choosing the hypermeters depends on how you want to control the output volume size of the convolution layer depending on your architecture. The output volume size of the Convolution layer depends on hyper parameter like the Input volume size, Kernel size, stride length and zero padding.

Mathematically, this can calculated using the formula,

Let say the input volume size = $W_1 \times H_1 \times D_1$

Then hyper parameter are defined as follows:
Let's say, Input volume size: $W_1$, Kernel size: F, Stride length: S and Zero padding: P.

Then the output size $W_2, H_2, D_2$ can be obtained by;
$W_2 = (W_1 - F + 2P) / S + 1$
$H_1 = (H_1 - F + 2P) / S + 1$
$D_2 = K$
The output size of the convolution layer will be $W_2 \times H_2 \times D_2$

### 3.4.1. Pooling Layer

Pooling layer is very usefully for reducing spatial size of the input and also the number of parameters from one convolution to another. Another advantage of using Pooling layer is that it controls overfitting. They are usually placed in between two consecutive convolution layers. Pooling layer does not care about the depth size of the previous layer, irrespective of the depth size of the input, it reduce the spatial dimension, using the MAX operator. The most common type of pooling utilises filter size of 2 and stride of 2 to reduce the spatial dimension of the input by half and discarding 75% of the activations. The Max operator will then select the max value from the 2 x 2 receptive field as shown in **Fig 16**. However, the depth dimension, will remain the same.

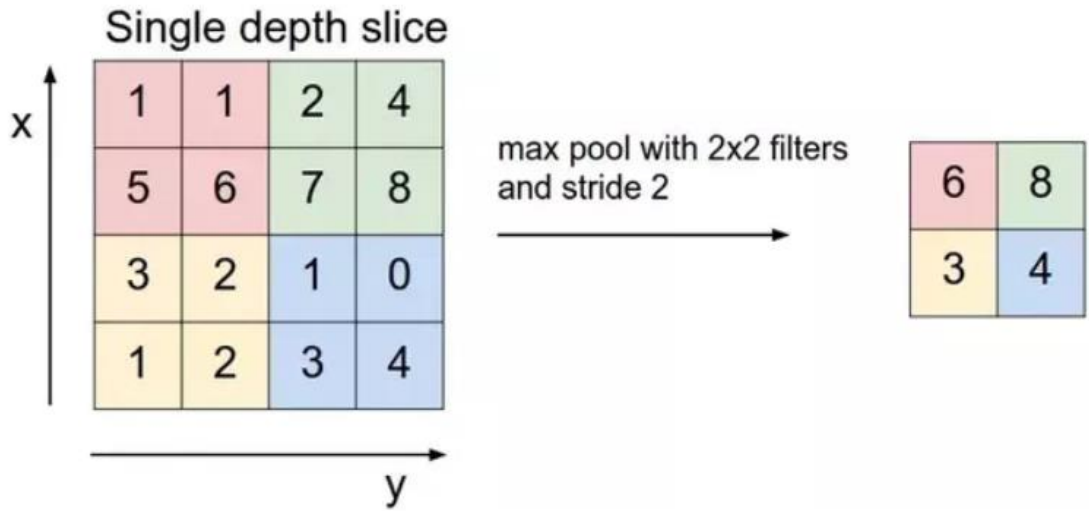

**Fig 16**. Max Pooling operation in action

### 3.4.3 Activation Function:

Activation Function introduces, non-linearity into the system, which is very useful in modelling complex relationship between features in higher dimensional space. Activation layer is made up of very simple nonlinear function, which takes some inputs and maps them to the range defined in the nonlinear function.

Most widely used Activation function in Deep learning are as follows.

**Sigmoid Activation function.**

The sigmoid function maps the all the input data into the range between 0 and 1. Mathematically, it can be expressed as.

$$\sigma(z) = \frac{1}{(1+e^{-z})} \qquad (2.0)$$

**Tanh Activation function.**

The tanh activation function maps all the data in the range between -1 and 1. Mathematically it can be expressed as;

$$\tanh(x) = 2\sigma(2x) - 1 \qquad (3.0)$$

**ReLU Activation layer.**

ReLU stands for Rectified Linear Unit, it is widely used in the deep learning community, this function clips the input real values z to 0 if z < 0 and allows all other real values.

$$R(z) = \max(0, z) \qquad (4.0)$$

**Leaky ReLU.**

Leaky ReLU function is slightly partial to negative values, consequently it allows small negative slope. Mathematically the function can be represented as,

$$f(y) = 1\,(y < 0\,)(ay) + 1\,(y >= 0)(y) \qquad (5.0)$$

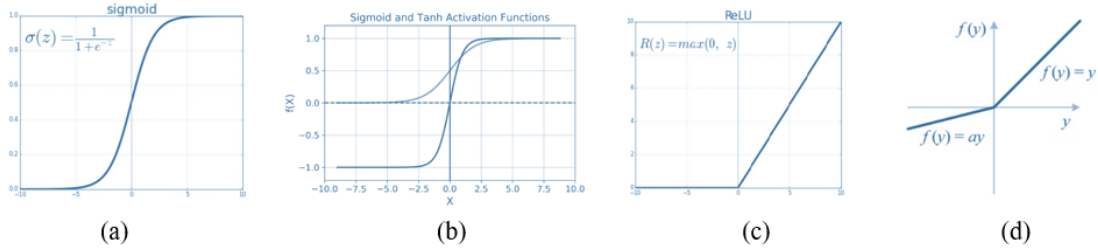

**Fig 17**. (a) Sigmoid activation layer, (b) TanH activation, (c) ReLU activation (d) Leaky ReLU activation

### 3.4.4 Fully Connected layer

A fully connected layer is layer that has a full connection to the previous layers neurons. As a result of which their activation can be computed as the weight sum of input from the previous layer and their inputs as shown in **Fig 14**.

The main difference between the Convolution layer and fully connected layer is that convolution layer are only connected to local region in the input, on the other hand in the fully connected network two adjacent layer are fully connected, however neuron within the same layer do not share any connection [26].

### 3.5 Loss function

One of the main goal of machine learning is optimisation of sets of weights that minimise the loss function. The Loss function that we are interested in this project for the Autoencoder is the Mean Squared Loss. Mean Square loss takes the square of difference between the input training example and its reconstructed output (reconstructed output) and then takes the Mean of all the sum of squared difference.

Mathematically, it can be expressed as, where $y_i$ is the input and $\tilde{y}_i$ is the reconstructed output

$$MSE = \frac{1}{n}\sum_{i=1}^{n} (y_i - \tilde{y}_i)^2 \qquad (6.0)$$

### 3.6 Optimisation and Gradient descent

Finding the set weight that minimise the loss function is non convex optimisation problem, they are many ways to find the set of weight that minimize loss function, instinctively one way to find the set of weights, would be to generate random weight and measure the accuracy of the model. However, it is not best way to approach this problem. Some ways to explore this problem are as follows[25 , 26],

**Iterative refinements.**

The core aim is to reduce the loss function, by finding the best set of weight parameter, a better way to achieve this is by using iterative refinement. It starts with a random weight *W* and then iteratively refines it, making it slightly better each time.

**Random local search.**

First, we assign a random set of weight and add a small perturbation to it and measure the loss of the given set of weights. If the loss for the given set weight and perturbation decrease, if continue exploring that direction.

**Gradient method.**

In the random local search, we randomly searched for the direction of the descent in the weight space that will help us finding the lower loss. However, using the Gradients of the loss function, we can implicitly compute the best direction of the steepest descent.

The gradient tells us the slope of the loss function along every dimension, this gradient then can be used later to update the weight of the model. Another important property of gradient is that it tells us the direction in which the function has the steepest rate of increase, however, it does not provide information about, how far the along the gradient should proceed along the given direction. Consequently, choosing the learning rate is very importance. Calculation of gradient of the loss function is a repetitive procedure to evaluate the model performance and update the model parameter (weights and bias)

**Calculation of gradient in the NN is done in two passes:**

**Forward Pass**

The forward pass computes the gradient values from input to output

**Backward Pass**

The backward pass is where back propagation takes, which begins at the end and recursively applies the chain rule to compute the gradients all way up to the input.

**Mini batch gradient descent.**

Often time, it is very wasteful to compute the loss for the entire training data to update a single parameter. Consequently, we only compute the gradient for a small subset of the training dataset called batches of the training data [25, 26].

## 3.7 Hyper parameter tuning

Setting the hyper parameter is a crucial design choice while building the testing the model , Often times machine learning practitioner try different values of Hyper parameter to get the best performance from the model architecture. The hyper parameter are only tuned or optimised during the training process, we do not tune the hyper parameter in the test set.

The best way to tune the hyper parameter is to split the training set into a slightly smaller subset called validation test. Use the validation set to find the best hyper parameter that works best on model

If the training data is small, we employ cross-validation, to tune the hyper parameter. Cross validation split the training dataset into a small number of chunks and randomly assign a single chunk as a validation set and remaining set as training dataset.

## 3.8 Data generation and preparation

**Data Pre-processing**

Data Pre-processing is most important factor in the feature engineering, which was discussed in section 3.3. With some domain knowledge and understanding of Neural Network architecture, the practitioner will have to train model to learn the select set of features without having to deal with overfitting or vanishing gradient problem [26]. Pre-processing is generally used if the dataset has different features at a different scales.

Some of the commonly used Data pre-processing techniques are Mean Subtraction, Standard Normalisation and Min-Max Normalisation.

**Mean Subtraction:**

Mean subtraction, subtracts the mean from all the features in a dataset, to which centres all the data around the origins

**Normalization:**

Normalisation is the process of scaling the all the dimension of the data to the same scale. One way to achieve this, is to divide the mean subtracted dataset by its standard deviation. Such that the dataset has a mean of zero and a unit variance.

$$z = \frac{x - u}{\sigma} \quad (7.0)$$

**Min Max Normalisation.**

The Min Max Normalisation scales all the features in the dataset to a minimum value and maximum value set by the user [26].

$$X_{norm} = \frac{X - X_{min}}{X_{max} - X_{min}} \quad (8.0)$$

## 3.9 Batch Normalisation

Batch Normalisation is a technique, which enables the loss of the NN model to converge faster during training. This is achieved by reducing the internal covariate shift in the layer. The role of the batch normalisation during training is to normalise the mini batch data by calculating the mean and standard deviation of data in that layer and normalise its values and scales and shift the resulting value to match the output layer size, to reduce the internal covariance of the layer for that given mini batch [27].

## 3.10 Weight Normalisation

Weight Normalisation is technique which parametrize the weight vector in the model. The parameterization is achieved by decoupling the length of the weight vector from their direction. Weight normalisation helps in improving the speed of convergence and conditioning the optimisation [28].

## 3.11 Dimensionality reduction.

Dimensionality reduction is core concept in the field of machine learning, it is generally used in dataset which has lots of dimension, which may impact the learnability and even may overfit the model. It always a good practice to understand your data. In short, dimensionality reduction is the process of reducing the number of random variables under consideration by obtaining a set of principal variables. By doing so, we are reducing the dimensionality of the feature space.

This can be achieved by following ways:

Feature Elimination: it is process of reducing the feature space by eliminating features which has little or no correlation

Feature Selection: we can compute the relative importance of certain features using variety of statistical test and select a subset of features in order of importance and eliminate the rest.

Feature Extraction: It is the process of extracting relevant features, often times we create a new independent feature, where each new independent feature is a combination of each of the old independent features.

There are two techniques to achieve this, namely Linear and nonlinear dimensionality reduction technique.

Linear Feature Extraction technique: Principal Component Analysis (PCA)

Non-Linear Feature reduction technique: T-SNE

### 3.11.1 Principal Component Analysis (PCA)

PCA is a linear feature extraction technique, it widely used to perform linear mapping of the data to a lower dimensional space to maximise the variance in the low dimension representation, and this is achieved by calculating the covariance matrix of the input data and finding its Eigen vectors. The Eigen vectors that corresponds to the largest eigenvalues are used to reconstruct a significant fraction of variance from the original data.

In other words, PCA interpolate the feature space in such a way that the least important features are dropped, while retaining the most valuable parts of all features [29].

### 3.11.2 T-SNE

T-Distribution Stochastic Neighbour Embedding (t-SNE) is a non-linear technique for dimensionality reduction that is especially used for visualisation of high dimensional data. It is designed to calculate the probability of similarity of data in high dimension space and the probability of similarity of the corresponding data in the low dimensional space.

It then tries to minimise the difference between these conditional probabilities (for similarities) in higher-dimensional and lower-dimensional space for a good representation of data in lower dimensional space [30].

In a nutshell, t-SNE minimizes the divergence between two distributions by measuring pair-by-pair similarities between the higher dimension and lower dimension.

## 4.0 CHAPTER 3: AUTOENCODER

Autoencoder is an unsupervised artificial neural network that learns to efficiently compress and encode data into a latent representation then learns to reconstruct the original data back from the latent representation. Autoencoder is has two connected networks called an encoder and decoder. An encoder network takes in an input, and converts it into smaller, compact representation, which the decoder network later uses to convert it back to the original input during inference.

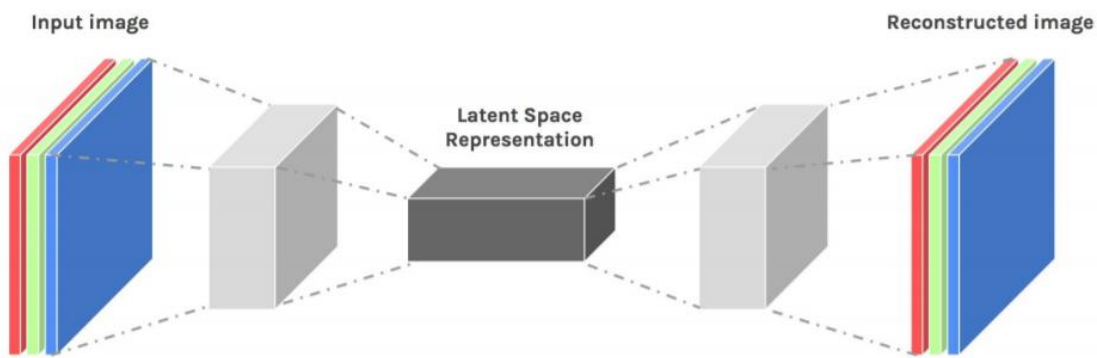

**Fig 18**, Typical convolutional Autoencoder.

Autoencoder is made up of the following components:

**Encoder:** Encoder section of the model takes in an input x and converts it to a much more compact representation called the latent vector z. The encoder must learn a nonlinear transformation to efficiently compress the data onto to lower dimensional space.

**Latent vector:** The latent vector is the compact representation of the input data. The vector holds all the information required for reconstruction. The latent vector only consists of important nonlinear representation in the low dimensional, which can be used for reconstruction.

**Decoder:** This section of the model learn to reconstruct the original data from the encoded latent representation, such that the reconstructed output is close to the original input as possible.

**Reconstruction loss:** The reconstruction loss is the measure of, how close or accurate the reconstructed data is to the original data. Generally Binary cross entropy loss or Mean Square Loss is employed to measure this loss and used in training to update the model parameters. This loss penalises the network for creating outputs that are very different from the input.

**Problems with Autoencoder:**

Vanilla Autoencoder are good enough for compact representation and reconstruction, However, they are very limited in functionality apart from few use cases like the demonising auto encoders.

The main problem with vanilla autoencoder, is that their latent space may not be continuous. If the latent space has discontinuities, then when we generate a random sample variation from there, then the decoder will generate an unrealistic output. Because the auto encoder has no data on how to reconstruct with that region of the latent space.

## 4.1 Variation Auto encoder

Variation Auto encoder (VAEs), is the widely used auto encoder because of its continuous latent space representation, which can be readily and randomly sampled for interpolation.

This is achieved by sampling two encoding vectors namely, the mean vector and standard deviation vector into a compact form as shown in Fig 19. The mean and standard deviation vector are sampled to Normal distribution to generate a continuous Latent vector representation Z.

Intuitively, the mean vector controls the where the encoding of an input should be cantered around, while the standard deviation controls the area on, how much from the mean the encoding can vary. The decoder learns from the distribution (rather than the single point on the latent space like in the vanilla auto encoder we discussed in the above section ) this allows the decoder to learn from the distribution which enables the decoder to generate random variation of reconstructed data .

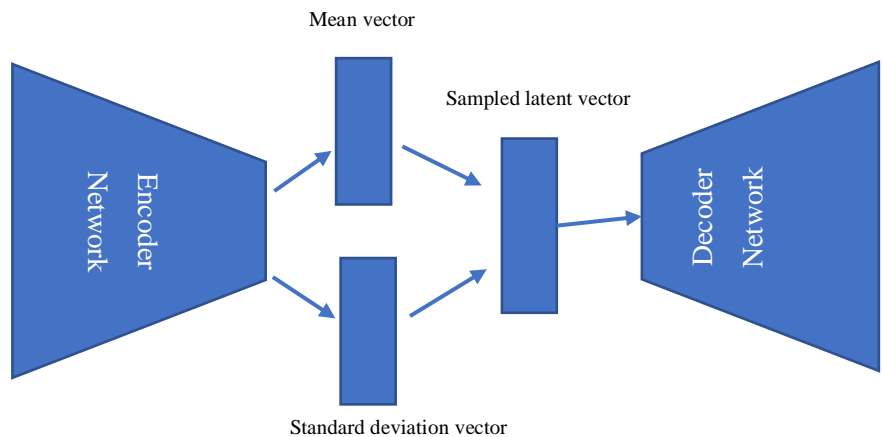

**Fig 19**. Abstracted Convolutional Variational Autoencoder

The model is capable to interpolate from the continuous latent space to generate random variation of input. But the resulting smooth latent space are clusters into group based on the shared characteristic of target output class within the dataset. Ideally, we want to decrease the inter class variability. This can be achieved by using Kullback-Leibler divergence (KL-divergence) in the loss function. The KL divergence between two probability distribution simply measures how much two similar probability diverge from each other. Intuitively this loss encourages the encoder to distribute all encoding evenly around the centre of the latent space [31].

**Mathematic Intuition,**

Let $X$ be the input data to the model, z be the latent Vector and $P(X)$ be the probability distribution of the data, $P(z)$ be the probability distribution of the latent variable and $P(X|z)$, be the distribution generating input data from the latent variable

The Objective here is to model the input data $P(X)$, using the law of probability, we can write it in relation with $z$ as follows:

$$P(X) = \int P(X|z)P(z)dz \qquad (9.0)$$

On further deduction we arrive at the objective function of the Variational Auto encoder VAE.

$$\log P(X) - D_{KL}[Q(z|X) \:||\: P(z|X)]$$
$$= E[\log P(X|z)] - D_{KL}[Q(z|X) \:||\: P(z)] \qquad (10.0)$$

Intuitively, the function conveys, to optimise the log likelihood of the *P(X)* under some encoding error. The encoder model $Q(z|X)$, probability of z latent vector given input data X, and the Decoder models $P(X|z)$, the probability of input data X given the latent vector z [31]. The Variational Autoencoder uses Variational inference to optimise two inference $P(X|z)$ (decoder) and $Q(z|X)$ (encoder) distribution using $D_{KL}$ KL divergence metric which minimise the difference between two distribution *P* and *Q*.

## 4.2 Conditional Variation Auto encoder CVAE

Conditional Variation auto encoder is an extension of variation auto encoder, it offer more control over the data generation. In CVAE models latent variable and data are both conditioned to some random variable.

On a Vanilla VAE, the objective is:

$$\log P(X) - D_{KL}\ [Q(z|X)\ ||\ P(z|X)] = E[logP(X|z)] - D_{KL}\ [Q(z|X)\ ||\ P(z)] \quad (11.0)$$

Objective function optimise the log likelihood of our data *P(X)* under some encoding error.

On examining the objective function of VAE, we can show that the VAE can't generate specific data from the training example. Because it encode the data *X* directly as into latent variable z. It does not understand the different class in the training data set. Because it does not take into account the label of the class. Similarly, the decoder part, only model the reconstructed output from the latent Z.

We could improve VAE by conditioning the encoder and decoder using the target label, let's say assume the target label as *c*.

We can condition the Objective function of the Variational Auto encoder (CVAE) with c for all of the distribution.

$$\log P(X|c) - D_{KL}\ [Q(z|X,c)\ ||\ P(z|X,c)] = E[logP(X|z,c)] - D_{KL}\ [Q(z|X,c)\ ||\ P(z|c)] \ (12.0)$$

So, the encoder is now conditioned to two variables *X* and *c*: $Q(z|X,c)$ and the decoder is also conditioned to two variables *z* and *c*: $P(X|z,c)$.

Now the encoder is conditioned with both the data and the target label to encode the latent vector with the condition. The latent variable is now distributed as conditional probability distribution (CPD) [31]. In practice, the CVAE could generate data with specific attributes, which cannot be replicated by VAE.

### 4.3 Using Auto Encoder to generate or interpolate 3D geometry

Recently many 3D mesh-based Auto encoder model were published for 3D reconstruction. These mesh-based Auto encoder model used either fully-connected or Graph CNN's based architecture to reconstructs 3D mesh or to perform some 3D shape analysis and also to learn the underlying deformation. Tan et al, introduced mesh auto encoder with rotation and translation in variative feature representation for the model to generalise to large scale deformation [22]. This model used fully connected architecture to adapt to the arbitrary connectivity in the mesh data structure. Tan et al also trained auto encoder model utilising graph convolutions to learn the localised deformation from the mesh [18]. Gao et al, proposed a transfer mesh deformation model by training a Generative Adversarial Network (GAN) with a cycle consistency loss to map shape in the latent space and variational mesh autoencoder learns to encode the deformation. Ranjan et al, developed Convolutional facial Mesh Autoencoder (CoMA) model mesh auto encoder for 3d facial reconstruction using a Spectral Graph convolution technique, by utilising a novel mesh down sampling and up sampling method [23].

## 4.4 Related Works.

### 4.4.1 Deep Appearance Models for Face Rendering

Modern VR hardware push the limit of immersive experience, by pushing the boundaries of computer graphics. However, driving the VR hardware at such high resolution require immense processing power at a high computational cost.

This paper aim to resolve the high computation cost, by creating a high fidelity facial model that can be built automatically from the camera capture set up and render the data real time at 90 Hz in VR.

This is a data driven model which rethinks rendering pipeline and learn the joints representation of facial mesh geometry and the multi view textures. The vertex position and its corresponding textures are modelled using a variational autoencoder. The latent representation produced is a compact representation of the non-linearity captured by the variational autoencoder. While capturing the non-linearity, the model also corrects any imperfect geometry present in the mesh [32].

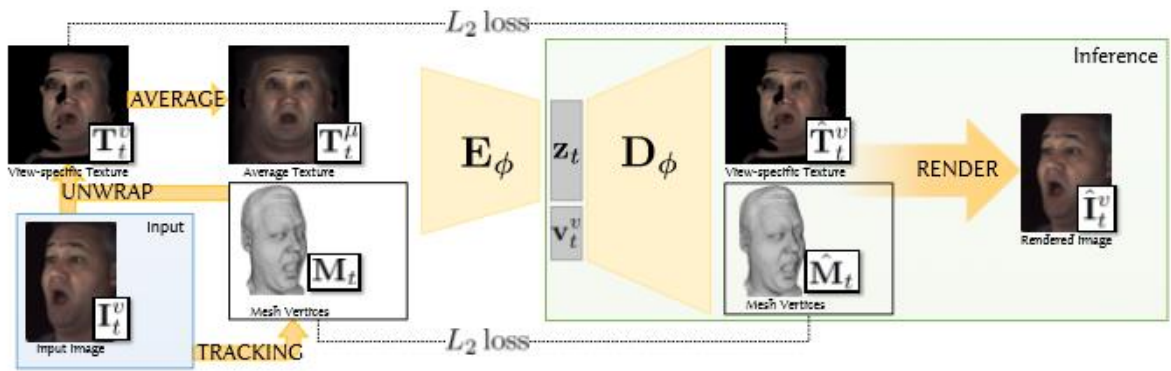

**Fig 20** , Abstract representation of the Deep appearance model for face rendering [32].

Architecture of the Deep Appearance Models for Face Rendering.

This autoencoder is built to learn and reconstruct the Texture and Mesh vertices. The encoder E learn to encode the learn representation into a compact latent vector Z and the decoder D learns to reconstruct the texture T and mesh sequence M. The model is conditioned at for different viewpoint representation of the texture, to enforce the canonical latent state over all view point. From the **Fig 20**, we can see that the auto encoder consist of two halves, the encoder E and decoder D. The encoder takes Mesh vertices M and texture and outputs the Gaussian distribution of the latent space,

$$E(T, M) \rightarrow \mu^z, log\sigma^z \qquad (13.0)$$

E is parameterised using the deep neural network with parameters at training, we sample from the Normal distribution to get $Z_t$

$$Z_t \sim \aleph(\mu^z, \sigma^z) \qquad (14.0)$$

And then pass it into the decoder D and compute the reconstruction loss.

$$D(Z_t) \rightarrow (T, M) \qquad (15.0)$$

The latent vector $Z_t$ is low dimensional nonlinear representation of the subject's facial state.

The Texture encoder is a convolutional neural network, which learn to encode the 3 channel 1024 x 1024 size input texture image to 256 vector size. Similarly the Mesh vertices is encoded into the using a fully connected artificial neural network, which also learn to encode the Mesh vertices to 256 vector representation. The 256 latent vector representation is concatenated into 512 vector representation, the $\mu^z$ and $log\sigma^z$ are then sampled from the latent representation to output a $Z_t$ latent vector [32].

The texture decoder consists of series of stride transposed convolution to increase the output resolution. On the other hand, the mesh decoder is made up of the full connect encoder which is designed to reconstruct the mesh to its original size.

The model utilises, the Leaky ReLU as its activation layer with 0.2 leakiness. The model is trained with Adam optimiser and L2 loss between the input texture and mesh sequence and the reconstructed mesh and texture sequence plus the KL-divergence between the prior distribution and the distribution of the latent space [32].

At testing and inference the decoder was able reconstruct both texture and mesh in around 5 milliseconds on NVIDIA GeForce GTX 1080 graphics card.

### 4.4.2 Variational Autoencoder for Deforming 3D Mesh Models

This paper studies and analyses the deformation 3D meshes using deep neural network. Deformation 3D meshes representation are flexible to represent 3D animation sequence and object with similar geometric shapes and connectivity. It is also invariant to large scale geometric deformation [22].

This paper also introduces a novel mesh representation which is rotation and translation invariant. The model is called Mesh Variation Auto encoder. Which explore the latent space behind the deforming 3D shapes and can generate new models which was not seen during the training process. The Mesh VAE is trained on 3D mesh which as same connectivity. Many existing datasets satisfy this property has 3d geometry in different pose and deformation. This work aims to produce a generative model, which can capture the large-scale deformation of the object. In order to achieve this model, it utilise a rotational and translational invariant input feature representation called RIMD (Rotation Invariant Mesh Difference), to effectively represent deformation using Variational auto encoder. The variational auto encoder is made is made up of fully connected network, along with a Mean Square Error loss [22].

### 4.4.3 Mesh Variational Auto encoders with Edge contraction Pooling.

This works proposes a new effective pooling technique, for graph-based convolutions, because existing graph based pooling technique limits the learning capacity of the network. In this paper a novel pooling technique is introduces to enhances the learning capacity of the network. Using this New Edge pooling technique developed in this paper, this variational autoencoder can learn with fewer parameter and perform better the Mesh VAE discussed earlier [33].

### 4.4.4 COMA

Generating 3D faces using Convolution Mesh Autoencoders.

This work focuses on generating a realistic facial reconstruction using Spectral Graph convolution layer and using a novel sampling technique call Mesh down sampling and Up sampling,

Traditional model learns the latent representation of the face using PCA, which is captures the linear transformation. Consequently, the deformation of 3d object is not captured. To overcome this COMA, has introduced a model which learns the non-linear representation of face using Spectral Convnets and a new sampling operation which captures the nonlinear representation. Which enables it to learn with less number of parameter and generates better reconstruction with lower error [23].

## 5.0 Chapter 4: COMPACT REPRESENTATION OF 4D VIDEO

This chapter is mainly focused on developing a Deep Neural Network model to generate compact and visually realistic representation of shape and appearance of the 3D model.

Compact representation of 4D video, is about how efficiently compressed the data is in terms of its space and time complexity. In addition to that we also explore the best way to encode and decode the compact representation of 4d video sequence. They are few ways to represent the mesh data structure in a compact way, like the popular dimensionality reduction technique called the Principal component analysis (PCA), it is also widely used is many applications for its dimensionality reduction capabilities and its ability to recover the original information. However, PCA only uses the linear transformation to reduce the dimensionality, which may not be the best choice when dealing with the non-Euclidean data structure like the mesh data structure. After the advancement of Machine learning and generative model like Autoencoder, VAE and GAN. Many researchers are pivoting towards deep learning architecture to effectively encode data using variants of autoencoder. Autoencoder are widely used to produce compact representation of data in a latent vector, which the models learn to encode during the training process. During training the model learns to encode the nonlinear transformation of the original data in a compact representation and low dimensional space called latent vector Z.

### 5.1 Data acquisition.

The primary source of 3D mesh training data are from CVSSP JP Mesh and texture dataset and Inria Multiview Thomas dataset and finally Body models datasets

INIRIA Multiview Thomas dataset consist of various human motion like Walking, climbing stair, left and right turn and dances and so on up to a 13 class labels [36].

BODY MODELS dataset is another popular dataset in the computer graphics community to learn shape and deformation of human motion, is made up of multiple male and female participants [34] with two class labels.

Finally, the third set of datasets used in the training for Mesh and Texture compact representation is the data from the CVSSP reconstruction pipeline. This dataset made up different genre of dance performance capture of a subject named JP with six class labels [35].

## 5.2 Data Analysis

In this data analysis, procedure we have tabulated the number of Training example per class label and also the number of class label categories in the dataset as shown in **table 1** . From the table 1, we can see that, the CVSSP JP and INRIA Thomas dataset has only around 200 – 250 training example per class. Which is a severe limitation. However, in the later experiment and evaluation we have shown a faithful reconstruction for both these datasets. We have also presented the reconstruction output for the model trained on Body models dataset, which performed really well and the reconstruction error was similar to the state of the art result obtained from the recent citation [22].

| Dataset | Class Label categories | No of examples across each label |
|---|---|---|
| CVSSP JP [35] | 6 | 250 |
| INRIA Thomas [36] | 13 | Approx. 150 -200 in each label |
| BODY MODELS [34] | 2 | 4800 |

**Table 1** , Dataset and no of training examples across class label categories

**PCA and TNSE analysis on the dataset**.

PCA and TNSE are techniques which are widely used to understand the dataset feature in low dimensional 2D space. This will help us understand how closely the two or more inter class target (class label categories of a dataset) features are clustered in the 2D space. In 2d or 3D low dimensional space we can easily tell that by looking at the cluster, if two or more cluster from different class label share common characteristics or similar pose (data attributes), then these data points will appear closer together. Similarly, we can tell if two or more class target are distinct, when they show a clear boundary of separation.

We have captured the low dimensional representation of the dataset to exactly see that how dataset with similar features or poses are arranged in the PCA and t-SNE low dimensional representation.

INRIA Thomas dataset has 13 label , they are represented as number for convenience in the table

| INRIA THOMAS | downstairs | jump | jumpforward | kick | left | leftsharp | low | push | right | rightsharp | run | upstairs | walk |
|---|---|---|---|---|---|---|---|---|---|---|---|---|---|
| | **0** | **1** | **2** | **3** | **4** | **5** | **6** | **7** | **8** | **9** | **10** | **11** | **12** |

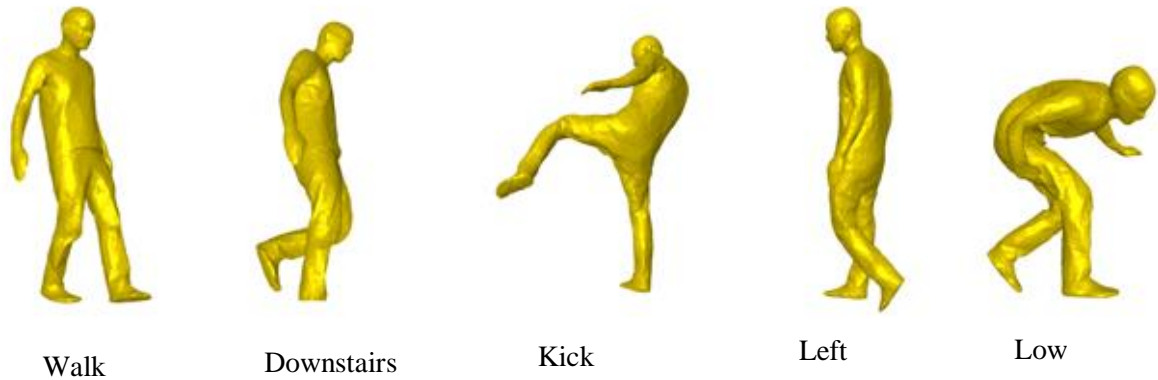

| Walk | Downstairs | Kick | Left | Low |

**Fig 21**. Thomas dataset target label and its corresponding mesh

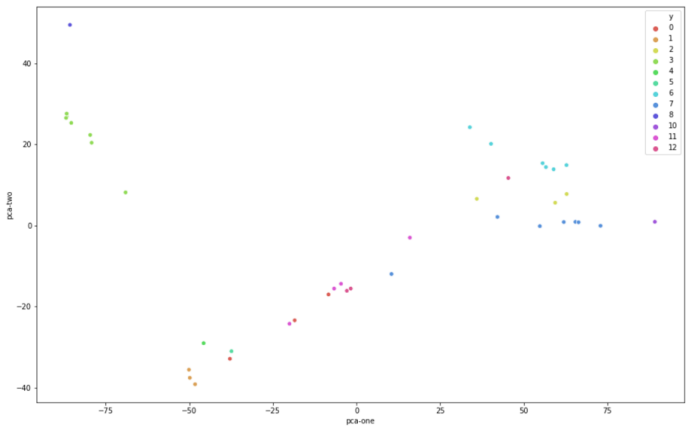 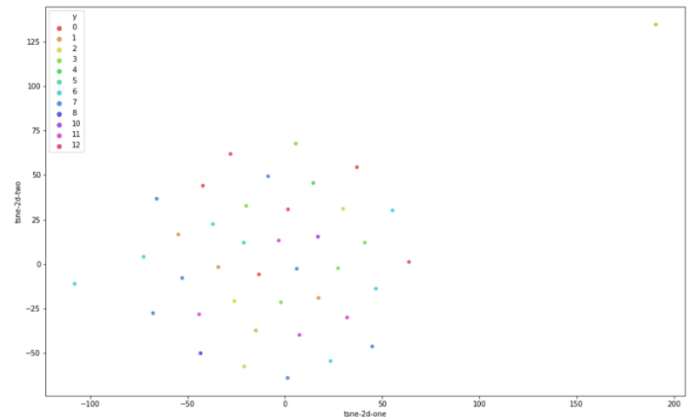

**Fig 22**. PCA 2D representation Thomas dataset      T-SNE 2D representations of Thomas Dataset

From fig 22 , and Fig 21 , we can see the Kick which is assigned to 3 ( light green in the PCA graph )  , in the PCA was separated from the other clusters . Similarly, we can see how clustered or separate a given label in the dataset in the 2d space. if they share common linear transformation data attributes those labels will be closely packed next to each other, for instance upstairs ( 11  in the PCA ) and walk ( 12 in the PCA) chart are aligned close to each other .   With this knowledge, it will enable us to improve our feature engineering and also designing our model .Similarly with T-SNE , we can visualise the  higher dimension data in the low dimensional space based on their probability of similarities between the high dimensional and low dimensional space.

## 5.3 Data processing.

The Inria Multiview Thomas and CVSSP JP mesh data structure values where not standardised, every class and subject had a different range of min and max values. Which is not ideal for Deep learning model. Because the features values in the dataset dictates the relative importance of that feature characteristic (due the weighted sum between the features values and weight). Therefore, it is mandatory to normalise the dataset within a standard range. This can be achieved by two ways , firstly , we can make sure the all the data in the training dataset has a Mean of zero and unit variance , this is achieved by calculating the mean value of the training dataset and then subtracting the mean with every value in the feature in the dataset and then dividing the resulting value by the standard deviation of the training dataset .

The second method is by using Min Max standardiser. This achieved by subtracting every feature by the Min value and then dividing the resultant value by the max value of the dataset. Both these methods are covered in section 3.8.

## 5.4 Creating custom dataset loader.

The all the Dataset used in the this project are custom dataset, which means all the training example must be labelled and loaded in sequence such that the both mesh and its corresponding texture sequence ( if texture sequence exists ) must loaded in unison in a mini batch fashion. Consequently, we have developed a custom data loader class, Steps for creating the custom data loader are as follows

1. Create a CSV file with the location of both Mesh and texture ( if texture exists)

2. Implement a Data transformation function to Pre-process the Data for normalisation

3. Create a Data loader class which outputs an iterator for the training dataset based on the transformation method and input dataset location specified in the csv file.

After Data acquisition, data cleaning and custom data class loader the next step is building the model from scratch, to effectively learn all the features in the dataset.

## 5.5 First Principle.

We use some of the ideals of first principle analysis to study the effect of changing various hypermeter and deduct higher level of abstraction on what happening inside the model. While, it is true that Deep Neural Architecture can model any function with enough training, Universal approximation theorem [37], but under the hood it is still a black box. So Hyper parameter tuning, and optimisation is still kind of an art, but not a solid science. So, we will use first principle technique to performance some experiments on the Model and make empirical deductions.

### 5.5.1 Problem definition that we aim to solve.

The aim is to explore different Deep learning technique to optimise the 4D video sequence for space and time complexity which can learn the compact representation of 4D video sequence and reconstruct it without affecting the shape and appearance.

### 5.6 Mesh based Autoencoder

The first model is the Mesh based Autoencoder: This Mesh based Auto encoder model learn to encode the latent representation and reconstruct the Mesh vertices. The Mesh based Autoencoder is built using a fully connected architecture, which enables the model to learn the arbitrary connection between the vertices. CNN is no the preferred choice because of the non-Euclidean nature of the mesh data structure and no proper local aggregation method is available for the mesh data structure in the vanilla Convolution Neural Network settings [22]. We will be studying and experimenting with Mesh based Variation Auto encoder and Conditional Variation Auto encoder.

### 5.7 Mesh and Texture based Autoencoder

The Second model is the Mesh and texture based Autoencoder: this model learns to jointly encode both the Mesh vertices ( 3643 x 3 ) and texture sequence (Image of resolution 1024 x 1024 x 3). The Mesh and texture based Auto encoder is built using two separate auto encoder architecture. The Texture based Auto encoder uses the Convolution Auto encoder architecture, which is well studied and yields very good reconstruction for Image, On the other hand the Mesh – based architecture uses fully connected architecture, similar to the first Model we discussed in the previous section.

This models jointly encodes both the texture and Mesh by concatenating the last layer of the encoding section of both Mesh based Auto encoder and Texture based Auto encoder. The joint concatenated representation is used to create the Mean vector and Standard deviation vector, which is for sampled Latent vector representation $z_t$. This joint Mesh and textured Latent representation are later used by the Decoder section to reconstruct the output Texture and Mesh Sequence.

Both the model discussed above is studied in two flavours of Auto encoder, namely Variational Autoencoder and Conditional variational Autoencoder setup.

The main difference in between these two variants of Autoencoder are

In Variational Autoencoder settings, the continuous latent space can be readily and randomly sampled for interpolation. Whereas on the other hand, Conditional Variational Autoencoder, offers more control over the data generation. In CVAE models both latent variable and target data are conditioned using one hot label. In practice, the CVAE could generate data with specific attributes, which cannot be replicated by VAE. Throughout the experiment we will be concatenating one hot encoded label to the input and reconstruction loss in case of CVAE.

## 6.0 Mesh Based Variational Autoencoder

The Mesh Auto encoder uses a full connected neural network architecture to effectively learn all the features and arbitrary connection in the mesh Data structure. We only use the 3D co-ordinated of the Mesh data structure as the input. Because the all the mesh in the data structure has consistent topological arrangements.

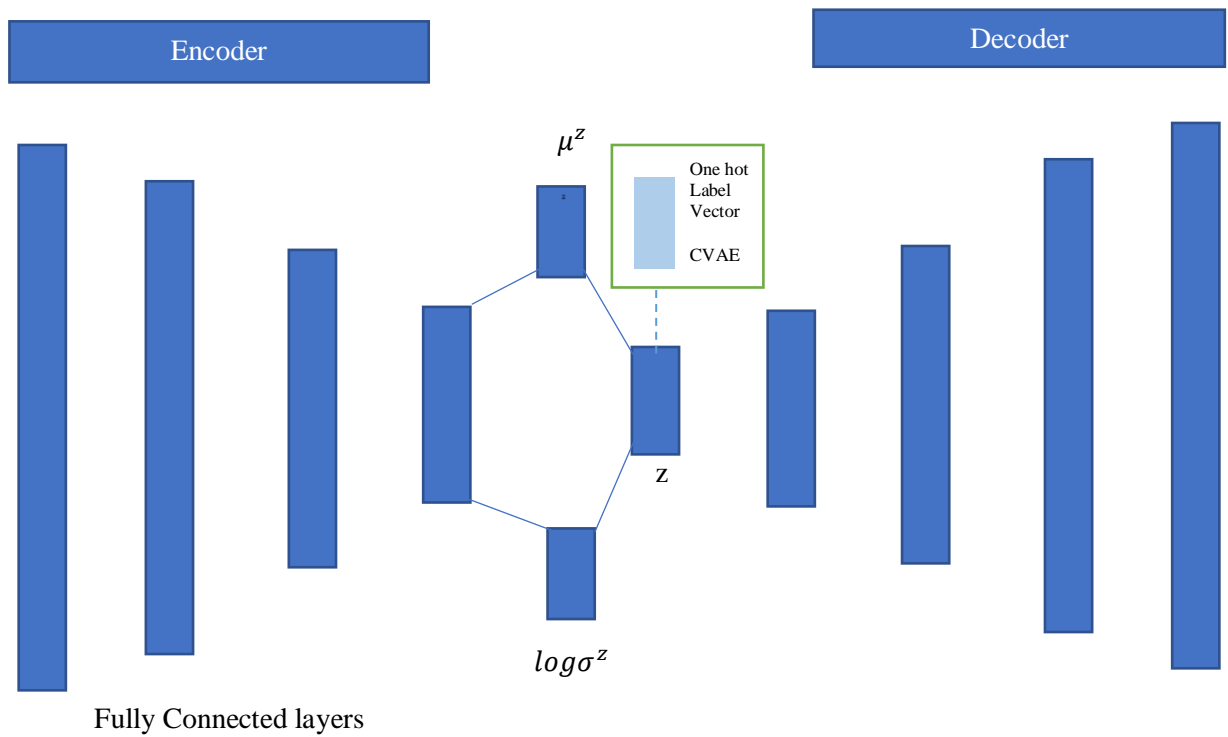

**Fig 23** , Fully Connected Mesh based Variation Autoencoder

The Input feature mesh data structure is mainly made up of Vertices and Faces. Vertices is N x 3 tensor, where N is the number of vertices and 3 is the dimension X,Y,Z in the three dimensional space. Faces is made up of three vertices, for a triangular mesh.

**Model Architecture**

Let us assume the flattened Mesh vertices size as M and the Latent Size as = L

| Encoder Layer | Decoder Layer |
| --- | --- |
| Linear (M, M/2) | Linear ( M/2 , M ) |
| Linear ( M/2, M/4) | Linear( M/4 , M/2) |
| Linear ( M/4 , M/8) | Linear ( M/8, M/4) |
| Linear( M/8 , M/12) | Linear ( M/12 , M/8 ) |
| Linear( M/12 , 256 ) | Linear(  256 , M/12 ) |
| Linear( 256, L ) | Linear( L, 256 ) |

**Table 2,** Mesh based Variational Autoencoder layer arrangements

The Flatten mesh representation is fed into the input layer of the fully connected network and consecutive layer in the fully connected VAE encoder is reduced by half for every hidden layer, until it reaches the latent vector dimension of 128 or 256, similar to Table 2 illustration. Usually the power of 2 is chosen as the latent vector size. The latent vector size can configure defending up of the application. In practice we try out different range of latent vector size and measure the average reconstruction error, to decide its size.

The decoder of the VAE doubles the size of the latent vector for every consecutive hidden layer, until it reaches the original size of the mesh.

In this Model architecture, Except for the final Nonlinear tanh layer in the Decoder part, every other Linear layer is followed by Batch Normalisation and Leaky ReLU , with a negative slope of 0.2 was used as a nonlinear activation layer, throughout the model. Final nonlinear activation on the decoder section, which produces the reconstructed output uses the tanh or sigmoid nonlinear activation layer, depending on the type of Data pre-processing technique employed .For instance,  if the Model is trained on Dataset which used Min Max normalisation , with a min value of -1 and maximum value of +1. Then in the decoder sector the last layer will use a tanh activation function, because it can represent the complete range of the Input data it was trained on. The Latent representation L is sampled and reparametrized with Mean and standard deviation vector .

This architecture of the model utilises Adam optimiser and Mean Square Error, to update the model parameters. We will also learn about the effect of using different hyper parameter and machine learning technique to improve the speed of the neural network. And analysis how it influences the models performance.

## 6.1 Training

Training is where the actual learning happens in the Deep Neural network. The network will learn the best set of model parameter (weights and bias) depending on the training dataset during the forward and Backward pass, with the help of a loss function ( MSE in our case) and a gradient descent algorithm ( Adam in our case). Depending on the loss output, the model parameter are updated during backward propagation, until the model learn the best set of weight and bias , in order for it to generalise well without overfitting or under fitting.

Generally, NN network are trained in Mini batch fashion, this is achieved by splitting the training dataset into small batches and training the model with that batch of dataset. Instead of training all the training dataset in one shot. Because, using all the training dataset sample is not effective in term the space and time complexity, plus it will consume a lot of GPU RAM. Consequently, training is usually performed on a small subset of the training data, called mini batch.

## 6.2 Model Hyper parameter and Batch size Selection.

Variational Beta ($\beta$) Variational Beta is a hyper parameter, which controls the representation capacity of the latent vector. Each variable in the latent representation is only sensitive to one single generative factor and relatively invariant to other factors. This disentangled representation is the most efficient representation. Therefore, a higher $\beta$ encourages a more efficient encoding.

Learning rate: it controls the extent of weight of the model is adjusted with respect to the loss gradient.

## 6.3 Experiments

Multiple Experiment where performed by maintaining main hyper parameter constant (learning rate, batch size and Normalisation) to understand the effect of changing few hyper parameters setting and seeing its effect on the reconstructed output to optimise the model.

Auto encoder has no label, it is Unsupervised learning problem, where the model learns the Identity function to reconstruct an output same as the original data it was trained on. The loss function we use throughout this experiment is Mean Square loss function, it measures the Euclidean distance (Reconstruction error) between the input data and reconstruction. Evaluation parameter we use are as follows,

**Average training reconstruction error** is the sum of all the training loss in a batch divided by the batch size.

**Average testing reconstruction error**, is the sum of all the test loss in a batch divided by the batch size

### 6.3.1 Experiment 1

Studying the VAE and CVAE

This Experiments study's the effect of using VAE and Conditioned VAE. The Conditioned VAE is condition with label in their reconstruction loss. The details about VAE and CVAE is discussed earlier in section.

| Model | Average training reconstruction error | Average testing reconstruction error. |
|---|---|---|
| **Mesh Based VAE** | 0.0618 | 0.00151 |
| **Mesh Based CVAE** | 0.0600 | 0.00131 |

**Table 3**, Mesh based VAE vs Mesh based CVAE

From the previous section we, know that the CVAE is conditioned with one hot encoded label in the input and the reconstruction loss, which gives more control over the data generation. This is also evident with the result we have got the fully connected Mesh Based CVAE as shown in **Table 3**.

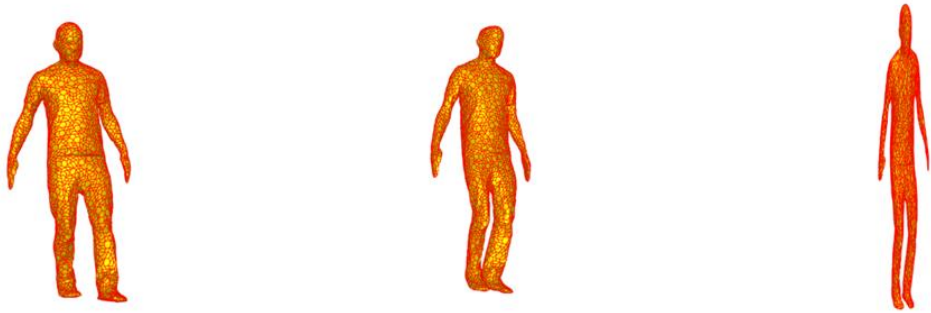

**Fig 24,** Original data: Thomas Downstair data    Reconstructed CVAE output    Reconstructed VAE output

From the **figure 24** , we can see that , the orignal thomas downstair data and reconstructred CVAE data set was able to capture the action and was able to render the mesh with structure and apperance , similar to the orginal mesh deformation. However, the reconstructed VAE output is failed to capture the structure and the output is poorly deformed when compared to the original data.

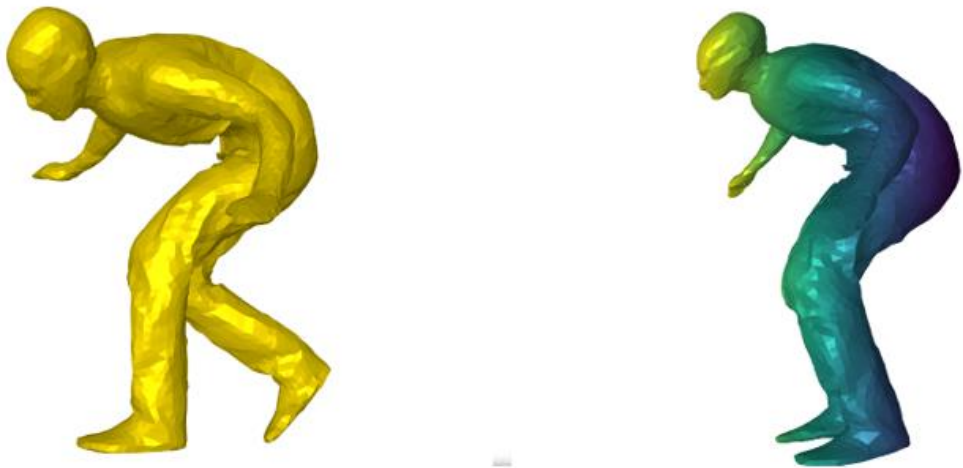

**Fig 25**.  Original data: Thomas low                     Reconstructed CVAE data Thomas low

From the above **figure 25**, we can see that , the conditional variational autoencoder was even able to generalise well to the certain large scale deformation. The orignal thomas low pose and reconstructed low pose almost have the same structure with some slight change in action . if we notice the right leg , we can clearly see the difference in action. Nevertheless, the conditional variational autoencoder was still able to generate a random sample , without altering the topology of the mesh structure and faithfully reproducing novel random samples by interpolating the latent space.

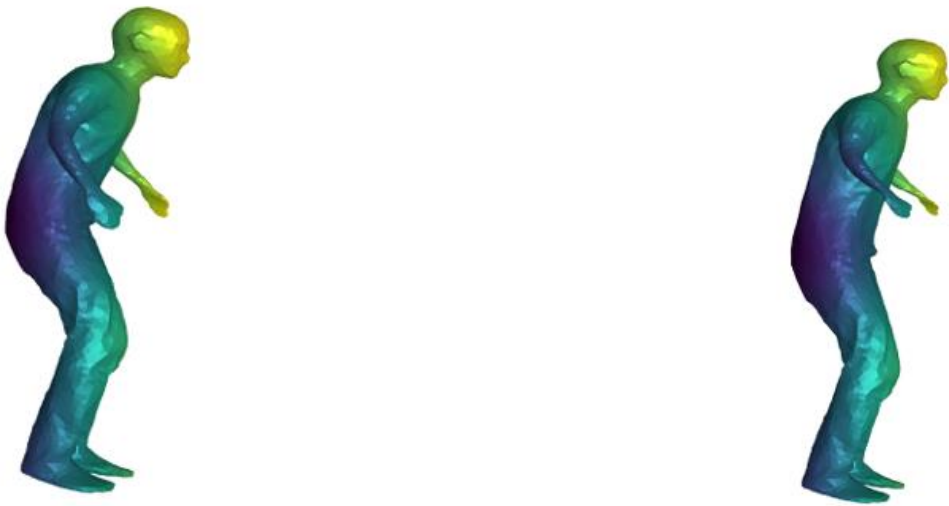

**Fig 26**, Original data: Thomas jumpforward          Reconstructed CVAE data Thomas Jumpforward

From the Fig 26, We can see that not all class of Thomas dataset was able to capture the large scale deformation, For instance, when we compare the hand of the original dataset with the reconstructed dataset , we can see that in the reconstructed data, both the hands of the model shows poor reconstruction and with very bad topology when compared to the original model.

### 6.3.2 Experiment 2

Studying the effect of changing the latent dimension.

The Selecting the best latent dimension is very important while designing an Auto encoder, because if the size of the latent vector is very small for the models training dataset. The model will under fit and consequently fails to learn the important features and characteristic in the dataset.

On the other hand, if the models latent vector size is too big, then the model will overfit and does not generalise very well in interpolation.

| Model | Latent vector size | Average training reconstruction error | Average testing reconstruction error. |
|---|---|---|---|
| **Mesh Based VAE** | 128 | 0.0618 | 0.00151 |
| **Mesh Based CVAE** | 128 | 0.0600 | 0.00131 |
| **Mesh Based VAE** | 64 | 0.061 | 0.00125 |
| **Mesh Based CVAE** | 64 | 0.0548 | 0.0019 |

**Table 4,** Result of Mesh Based VAE and CVAE with different latent vector size.

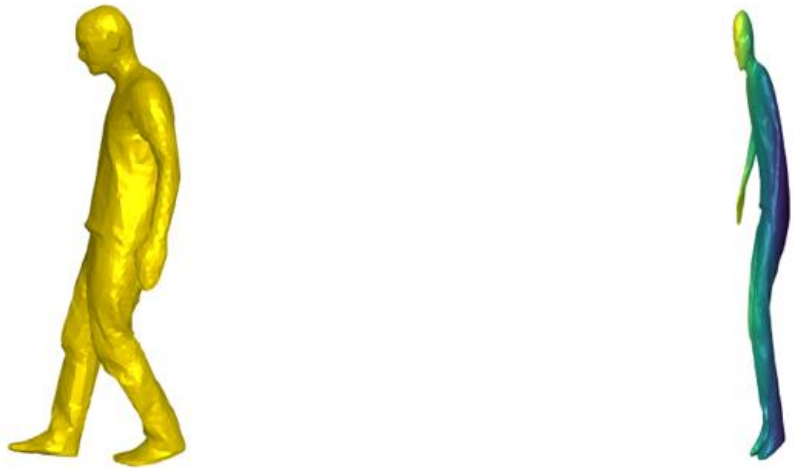

**Fig 27** , Original data: Thomas          Reconstructed CVAE data Thomas  with 64 latent size

From the **Fig 27**, we can see that, the original model and reconstructed CVAE data with latent representation of 64 vector. It is evident the latent vector representation of 128 was able to faithfully reconstruct the output of the topology of the original data as shown in **Fig 25 , Fig 24**  , however the latent vector representation 64 failed to capture or learn the similarly topology, this is because the representation capability of the model with latent vector 64 is limited and adds many constraints for the model to encode in the 64 length vector, as a result of which the model under fits .

6.4 Evaluation.

From the previous experiments we can see that Mesh Based CVAE network with a latent representation of 128 vector length, produced a faithful and topologically similar reconstruction when compared to the original mesh data. Although, it not quite visible from the **Fig 28.** We can clearly see the reconstruction mesh data from **Fig 24** and **Fig 25** .

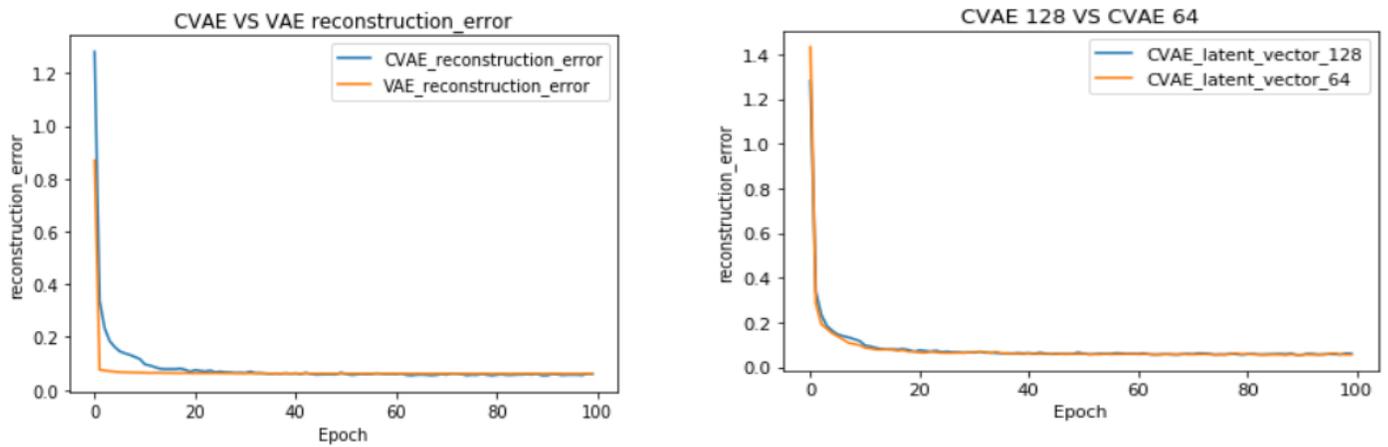

**Fig 28**, Graph showing the reconstruction error of CVAE vs VAE and   CVAE latent 128 vs CVAE latent 64

6.5 Inference of Decoder.

After the network is trained, we have just used the Decoder Section of the Mesh-based Autoencoder for inference, to generate the reconstruction from random latent vector interpolation.  During inference, the model will be effectively use the all the best set of parameters that it has learned during the training process. and also uses the detangled representation from the low dimension to interpolate from the random  latent vector to reconstruct the mesh geometry.

Random Latent interpolation during inference on the INRIA Thomas dataset

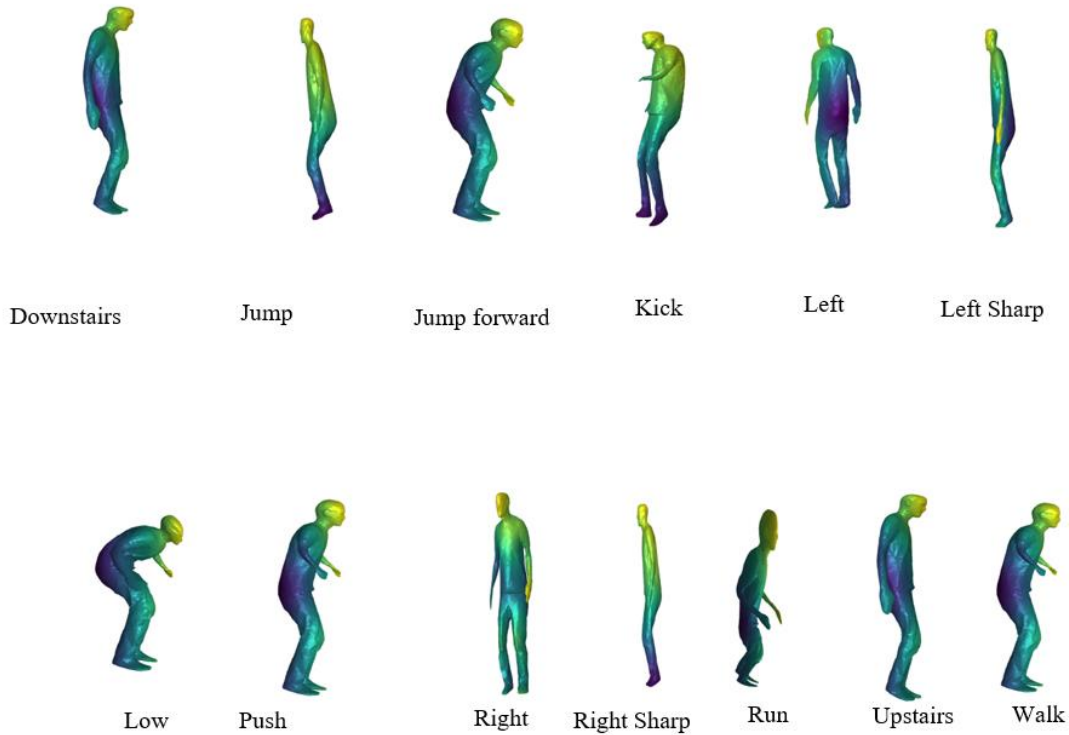

Figure above is the Random latent interpolation of all the class label during inference. By checking the latent interpolation for each class, we get to see, which class label was effectively encoded in the compact representation and can help in designing a better model.

## 6.6 Comparison with state-of-the-art Method

The Body Models dataset is widely used in the computer graphics community for shape analysis. Therefore, in this section we can compare some recent work which has used autoencoder for 3D reconstruction, we compare the performance of our model and other recent citation which claims to have produced state of the art result like the variational Mesh auto encoder [22] using this standard dataset.

| Model | Average training reconstruction error | Average testing reconstruction error. |
|---|---|---|
| **Mesh Based CVAE** | 0.0025 | 0.001512 |

**Table 5** , Reconstruction error for Body Model dataset.

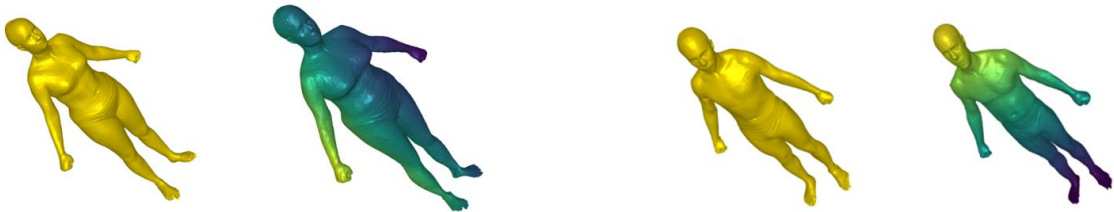

**Fig 29**, Output of Original Body model (yellow mesh ) vs Reconstructed output (Gradient of blue)

From Fig 29 and Table 5 we can see that quality of the mesh is reconstruction is greatly improved and the reconstruction error is also dropped significantly. The reason being the Body model dataset has a training example of 4800 for each class, when compared to the CVSSP JP and INIRIA Thomas dataset. Another important reason is that Body model uses very dense vertices ( 12500 x 3 ) 3d coordinated to provide such level of detail in the mesh structure.

On comparing the result of variational Mesh Autoencoder using 3D coordinates produced $1.57 \times 10^{-3}$ as reconstruction error. Our Mesh based model also yields very similar result with reconstruction error of $1.51 \times 10^{-3}$ on a similar dataset.

## 6.7 Mesh rendering time during inference

The primary aim of the project is to test the feasibility of rendering the 4D video sequence in XR headset like the Microsoft HoloLens. Generally, we use 24 frame per second to create the illusion of motion using the image in a generic video. Similarly, we can create an illusion of motion by rendering the temporally coherent 3d model at a desired frequency or refresh rate. In most VR application, the refresh rate is around 60 Hz , to deliver the best experience for the user.

So, we need to test, if the model during the inference is capable of rendering at the refresh rate, to be able to successfully render them in XR platform like the Microsoft HoloLens at the refresh rate of 60 Hz or higher.

On testing the model at inference, the model took approximately 0.00679 seconds to render a single mesh. This 0.00679 second also includes the times taken by the decoder to reshape and transfer the 3d coordinate to CPU. Therefore, we can easily render the 3d mesh at refresh rate even greater than 90 Hz, so it is indeed feasible to render complex mesh at this refresh rate, Similar experiment was also setup by in [32] and achieved 5 millisecond time to render the 3D mesh.

## 7.0 Mesh-texture based Auto encoder

This model learns to jointly encode both the Mesh vertices ( 3643 x 3 ) and texture sequence (Image of resolution 1024 x 1024 x 3). The Mesh and texture based Autoencoder is built using two separate auto encoder architecture. The Texture based Auto encoder uses the Convolution Auto encoder architecture, which is well studied and yields very good reconstruction for Image, On the other hand the Mesh – based architecture uses fully connected architecture, similar to the Mesh based Autoencoder we discussed in the previous section.

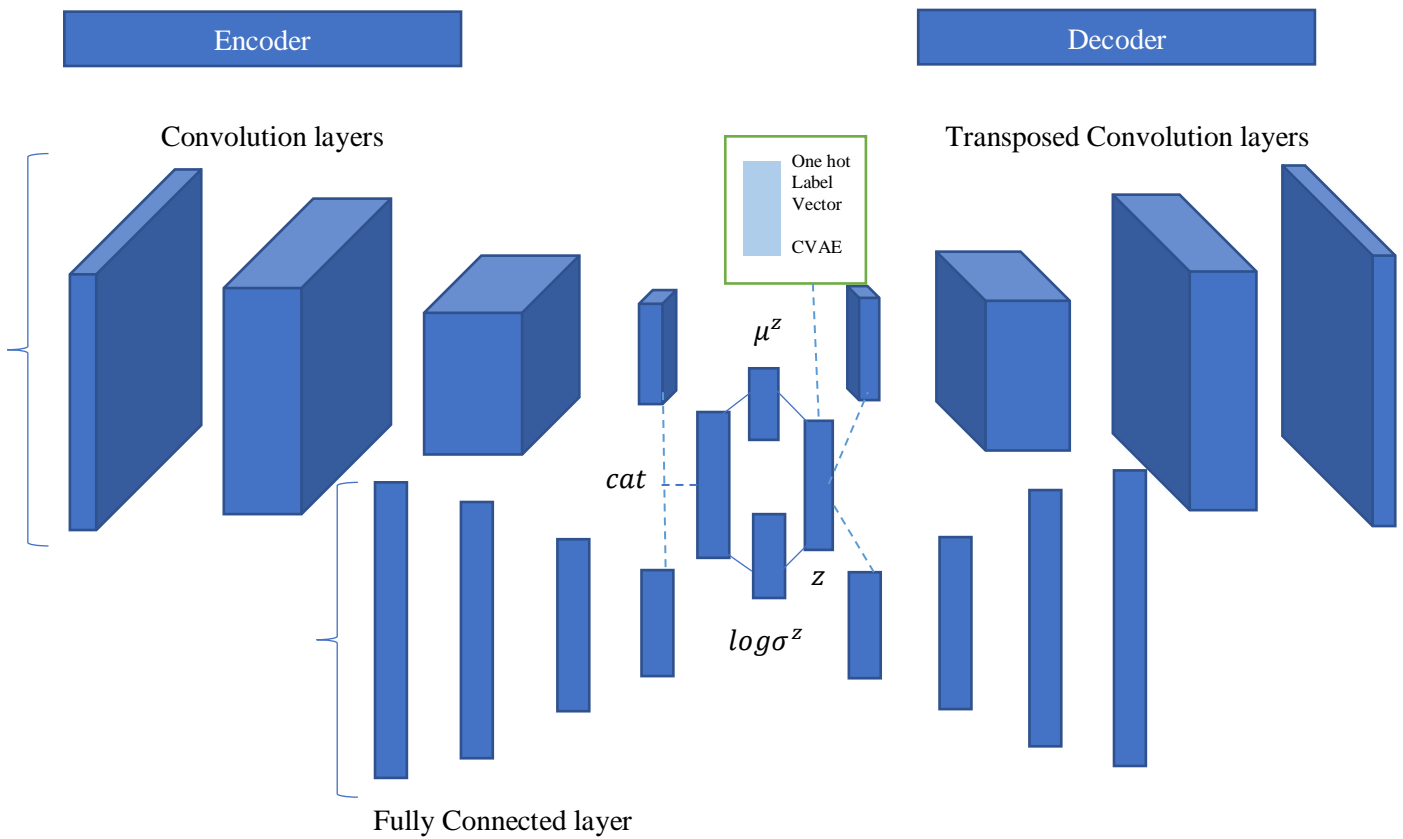

**Fig 30** , Mesh and Texture based Variational Autoencoder

## 7.1 Model Architecture

The Mesh and texture-based Auto encoder is built using two separate auto encoder architecture. The Texture is trained using Convolution Auto encoder architecture as show in **Fig 30**, which is well studied and yields very good reconstruction for Image, On the other hand the Mesh – based architecture uses fully connected architecture,

If we were to remember the second objective of this project, it was to produce a compact representation of the both mesh and texture sequence for a 4D video sequence. This model aims to fufill the second objective by jointly encoding both mesh and texture sequence.

This model is trained on a texture sequence of dimension (1024 x 1024 x 3) and Mesh sequence of dimension of (~ 4000 * 3 ) in total if we flatten this representation as a single vector then it equates to = 3145728 + 12000 = 3157728 of vector length . In this Model we will be encoding the them into a latent vector representation of 256 in length. From 3157728 to 256 length latent vector representation is roughly $8 \times 10^{-3}$ reduction in size.

**Convolution Encoder layer**

**Transpose Convolutional Decoder Layer**

| Layer | Input Size | Output Size | Layer | Input Size | Output Size |
|---|---|---|---|---|---|
| **Conv1** | 1024 x 1024 x 3 | 512 x 512 x 16 | **TransConv8** | 512 x 512 x 16 | 1024 x 1024 x 3 |
| **MaxPool1** | 512 x 512 x 16 | 256 x 256 x 16 | **TransConv7** | 256 x 256 x 16 | 512 x 512 x 16 |
| **Conv2** | 256 x 256 x 16 | 128 x 128 x 32 | **TransConv6** | 128 x 128 x 32 | 256 x 256 x 16 |
| **MaxPool2** | 128 x 128 x 32 | 64 x 64 x 32 | **TransConv5** | 64 x 64 x 32 | 128 x 128 x 32 |
| **Conv3** | 64 x 64 x 32 | 32 x 32 x 64 | **TransConv4** | 32 x 32 x 64 | 64 x 64 x 32 |
| **MaxPool3** | 32 x 32 x 64 | 16 x 16 x 64 | **TransConv3** | 16 x 16 x 64 | 32 x 32 x 64 |
| **Conv4** | 16 x 16 x 64 | 8 x 8 x 128 | **TransConv2** | 8 x 8 x 128 | 16 x 16 x 64 |
| **MaxPool4** | 8 x 8 x 128 | 4 x 4 x 128 | **TransConv1** | 4 x 4 x 128 | 8 x 8 x 128 |
| **Linear1** | 4 x 4 x 128 | 1024 | **Linear1** | 1024 | 2048 |
| **Linear2** | 1024 | 512 | **Linear2** | 512 | 1024 |
| **Linear3** | 512 | 256 | **Linear3** | 256 | 512 |

**Table 6** , Layer used in Convolutional Variational Texture based Variational Autoencoder.

From **table 6** , we can see that the Convolutional Encoder layer input a 1024 x 1024 x 3 image and follows a certain pattern in reducing the size of the convolution layer in the first convolution layer the spatial dimension ( W x H ) of the input image is reduced by half and the depth of the is increased 16 , following the first convolution layer the spatial dimension ( W x H ) continues to reduce by half but the depth seem to double after every two consecutive layer. This is because Maxpooling layer between every Convolution layer only reduces the spatial dimension but the depth remain the same .

This pattern continues until the output tensor of size 4x 4 x 128 , from this point onwards the tensor is flattened and fed to linear layer , the three consecutive linear layer reduces the flatten vector to the vector of size 256, which is the last layer of the encoding layer .

The Fully connected Mesh based Autoencoder uses the same architecture as discussed in **section 6.0** . The Model Architecture layer is similar to that of the layer shown in **table 2** .

This models jointly encodes both the texture and Mesh by concatenating the last layer of the encoding section of both Mesh based Auto encoder and Texture based Auto encoder. The last encoding layer of both this model Architecture is concatenated and the joint concatenated representation is used to create the Mean vector and Standard deviation vector, which used for sampled Latent vector representation $z_t$. . This joint Mesh and textured Latent representation is later used by the Decoder section to reconstruct the output Texture and Mesh Sequence.

Similar to Encoder Convolution section, the transpose convolution section follows the same pursuit , in the opposite direction ,it first start from the first three linear layer in Decoder section , and once it reach the tensor od 2048 size , the tensor reshape to 4 x 4 x 128 and continues Up scaling the them in similar fashion until the decoder reaches the original 1024 x 1024 x 3 image dimension.

Like all the model architecture we have discussed , the variational autoencoder trained , can see be converted to conditional Autoencoder by concatenation the one hot label vector to the input and reconstruction loss , to yield lower reconstruction error , detailed discussion on this topic is discussed on section 4.2 .

## 7.2 Dataset used

For this model the only dataset that we can use is the CVSSP JP Dancing dataset, this dataset contains both Mesh and texture sequence required for training the model Mesh and texture based Auto encoder. Before training we need to pre-process the dataset.

**Data pre-processing**

For multi modal model with different data feature representation, it necessary to make sure that Image and Mesh dataset are represented in a complimentary way. This enables the models to solve more complex task and improve the performance of the model. When the data feature of both the image and mesh are represented in the standardised format, then model will give equally importance in encoding characteristic of both the Mesh and Image data representation in the latent vector. In this Mesh-texture based Variational Autoencoder network, we have employed Min Max normalisation for both the Image and Mesh, with the min value of -1 and max value of +1. This enables us to use a tanh nonlinear activation to faithfully reconstruct the output.

## 7.3 Training.

Training the multi-modal model, is very difficult, we need to make the data features are represented in a similar manner, therefore we use min max pre-processing with the same range for both the texture and Mesh sequence. From the figure 30 , we can see that texture sequence is trained on the Convolutional based Auto encoder, on the other hand the Mesh is trained on the fully connected neural network architecture. In order for the model to jointly encode both the texture and mesh sequence, we concatenate the last layer of the encoder and produce two encoding namely the mean vector encoder and standard deviation vector encoding and sample the latent vector $Z_t$ from the mean and standard deviation. This architecture is similar to the Deep Appearance model for face render by Facebook [32], except for the fact that , they learn the encoding of facial topology , which does not have large scale deformation like the JP dataset from the CVSSP .

Similar to the Mesh based Auto encoder we discussed earlier in section 6.0 , this model is trained in minbatch fashion , the Optimiser of choice is the Adam Optimiser , with a learning rate of .001 .

The Loss in the Mesh – Texture based Auto encoder is a combinational of Mesh Loss and Texture Loss . Mean Square Error loss function is the preferred choice in our model , from which we can calculate the Euclidean distance for the reconstruction error.

For the Texture based Convolutional Auto encoder section, we calculate the L2 loss between the original texture data ($T$) and reconstructed texture data ($\tilde{T}$). Mathematically,

$$MSE(T) = \frac{1}{n}\sum_{i=1}^{n} \left(T_i - \tilde{T}_i\right)^2 \tag{16.0}$$

For Mesh based Fully connected autoencoder, we calculated the L2 loss between original Mesh data ($M$) and reconstructed output ($\tilde{M}$). Like the Mesh-based Autoencoder discussed in the previous section.

$$MSE(M) = \frac{1}{n}\sum_{i=1}^{n} \left(M_i - \tilde{M}_i\right)^2 \tag{17.0}$$

The overall custom loss function is the L2 loss between the input texture and mesh sequence and the reconstructed mesh and texture sequence plus the KL-divergence between the prior distribution and the distribution of the latent space.

This can be mathematically represented as,

$$Loss = MSE(T) + MSE(M) + \beta * KL\ loss \tag{18.0}$$

Where, $MSE(T)$ is the L2 loss of the texture sequence and $MSE(M)$ is the L2 loss of the mesh sequence and $\beta$ (variation beta) is a hyper parameter and $KL$ is the KL divergence, which measure the difference between the prior distribution and distribution in the latent space, see section 4.1 about variational auto encoder.

### 7.4 Experiments

Like the last Mesh based Auto encoder we will perform multiple experiment to understand the effect of the changing various hyper parameter settings.

The loss function is the custom loss function which is based on Mean Square loss function, it measures the Euclidean distance (Reconstruction error) between the input data and reconstruction.

We will be measuring the Average training reconstruction loss and testing loss to evaluate the model performance.

Average training loss reconstruction error is the sum of all the training loss in a batch divided by the batch size.

Similarly, Average testing loss reconstruction error, is the sum of all the test loss in a batch divided by the batch size

### 7.4.1 Experiment 1

Studying the Mesh texture Based VAE and CVAE.

This experiment studies the influence on reconstruction error and reconstructed output based on choosing a VAE and CVAE for the Mesh Texture Based Auto encoder. We know for the previous section the advantage of choosing a CVAE over VAE. Nevertheless, let's put this into test.

| Model | Average training reconstruction error | Average testing reconstruction error. |
|---|---|---|
| **Mesh-Texture Based VAE** | 0.13 | 0.0104 |
| **Mesh-Texture based CVAE** | 0.11 | 0.00652 |

**Table 7**, Reconstruction error for Mesh_ texture based VAE vs CVAE

From the table 7, we can see that the CVAE training and testing reconstruction error is lower that the VAE.

From the previous section 6.3.1 we, know that the CVAE which is conditioned with one hot encoded label in the input and the reconstruction loss, will give more control over the data generation. This is also evident with the result we have got the fully connected Mesh Based CVAE.

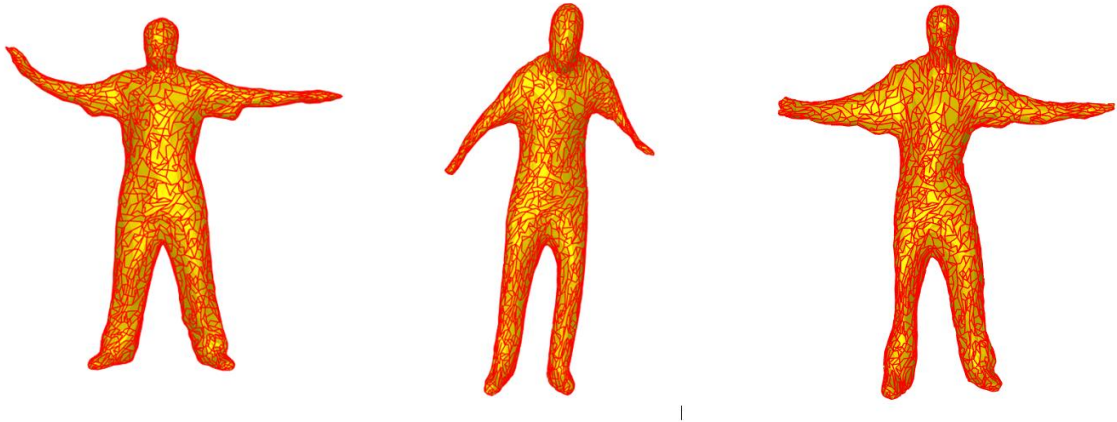

**Fig 31** , Original data: JP pop     Reconstructed data VAE: JP pop     Reconstructed data CVAE: JP pop

For the Figure 31, we can see that the reconstruction of the CVSSP JP pop model is affected by the large-scale deformation and as result both the arms of the reconstructed data is not well defined in term of shape and appearance when compared to the original data. However the reconstruction output of CVAE is much better in term of preserving the structure and appearance.

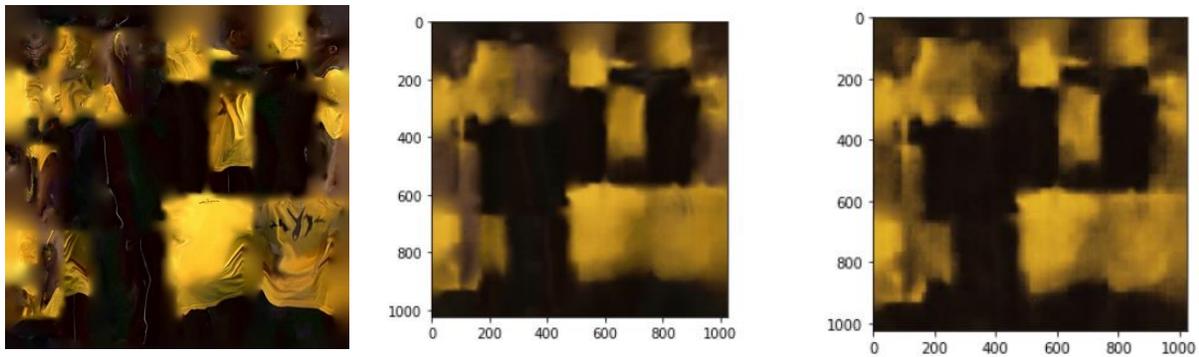

**Fig 32**. Original data: JP     Reconstructed data VAE:JP     Reconstructed data CVAE:JP

From the above **figure 32**, we can see that the original dataset Texture image and reconstruction image of both VAE and CVAE. The model is still able to successfully decode both the texture and mesh sequence. However both the VAE and CVAE reconstruction images where bit blurry , we believe this is due to the the number of training example available for for this dataset , because there is only around 1000 images in the entire dataset for a complex model with 1024 x 1024 x 3 image resolution. For comparison the celeb100 dataset has a image resolution of only 128 x 128 x 3 dimension and has a training example of 200K in total.

### 7.4.2 Experiment 2

Studying the effect of changing the latent dimension

The selecting the length of the latent vector is very crucial effectively encoding the lower dimensional representation of the both texture and Mesh Sequence. If the latent vector is too big then the model will try to over fit and if the latent vector is too small, then the model will try to under fit.

| **Model** | **Latent dimension** | **Average training reconstruction error** | **Average testing reconstruction error.** |
|---|---|---|---|
| **Mesh-Texture Based VAE** | 512 | 0.13 | 0.0104 |
| **Mesh-Texture based CVAE** | 512 | 0.11 | 0.00652 |
| **Mesh-Texture Based CVAE** | 256 | 0.106 | 0.00606 |

**Table 8**, Reconstruction error for Mesh-texture CVAE _ 512   CVAE _ 256 latent dimension

From the table 8, we can see that 256 latent vector representation produce a better average training reconstruction error when compared to the 512 latent vector representation for the Mesh-Texture based CVAE model.

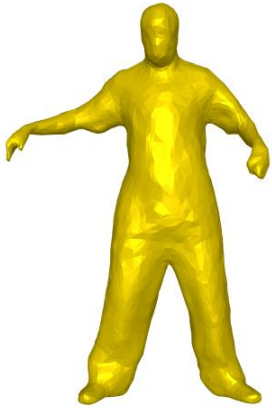 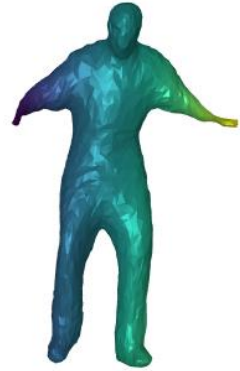

**Fig 33,** Original data: JP                              Reconstructed CVAE data JP : Latent vector 256

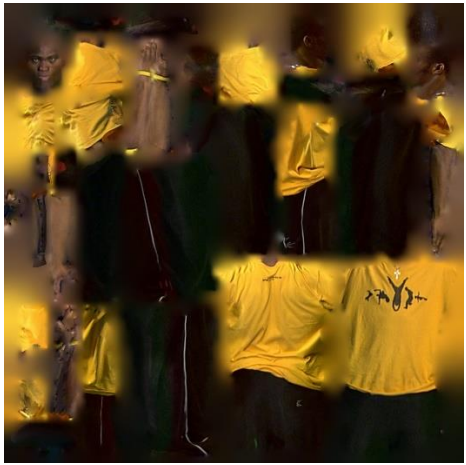 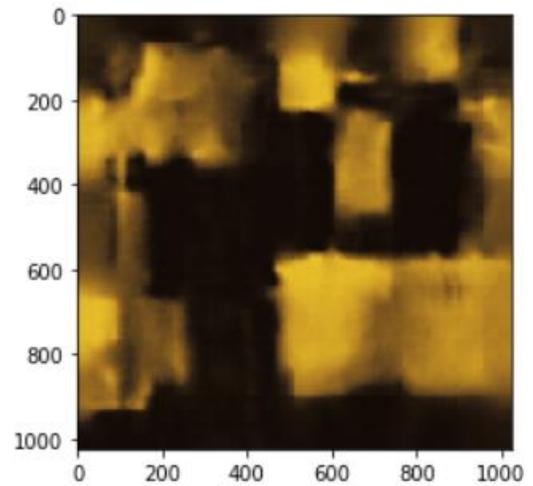

**Fig 34,** Original texture data: JP                     Reconstructed texture image CVAE_256: JP

From the figure 34 we can see that, there is little blurriness in the reconstructed output tand original image. Moreover, from the table 8, we can see the Mesh and Texture based Auto encoder with latent vector 256 performed relatively better than the 512 counterparts. Consequently, we can infer that, 256 latent vector representation can effectively encode both Mesh and Texture sequence.

## 7.5 Evaluation.

From the table 8, can see that, the Mesh and texture based CVAE with the latent dimension vector of size 256 has a lower reconstruction loss for both the training and testing. Also From figure 33 and 34 we can see that the Mesh and texture based VAE did not quite capture the geometry of the original data. Also the reconstruction of the image output, is blurry. But still captures most of the data attributes from the original image. This reason can be attributed to the limited number of training sample in the data. In CVSSP JP dataset, each label has only around 150 examples. Which makes it difficult, for the network to capture all the data attributes for both the Mesh and texture. This model can generalise better, with more data and produce a even better reconstruction output for both Mesh and texture. From figure 35, we can also show and summarised the result of the experiment performed.

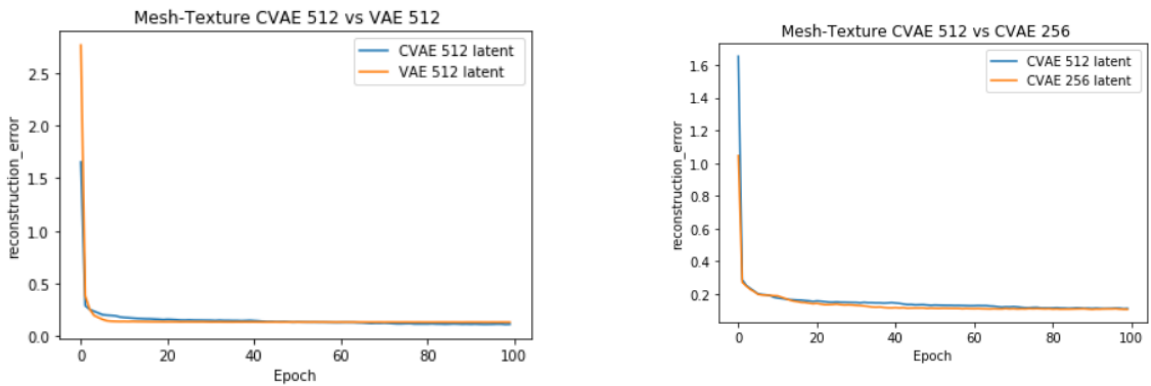

**Fig 35,** Reconstruction error graph for all the experiment performed.

## 7.6 Model Inference.

We know that, during inference the network generates outputs reconstruction using only the decoder section of the Autoencoder, which the network learned during training process. But in the case of Mesh-texture based Autoencoder, the network has jointly encoded both the Mesh and texture in the latent representation. So, during inference, the decoder section of the Mesh-Texture based Autoencoder will yield both the Mesh and texture reconstruction output. Both the output is shown in the figure 36 for all the class label of the mesh and in figure 37 reconstructed texture images are generated.

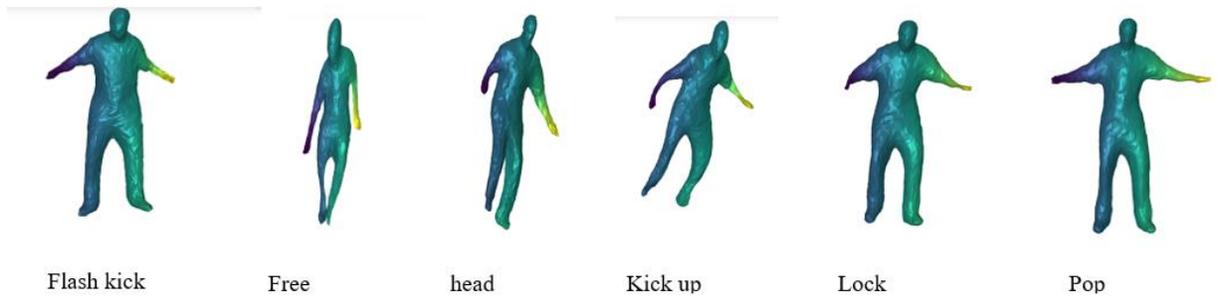

**Fig 36**, Random Latent interpolation of mesh during inference on the CVSSP JP dataset

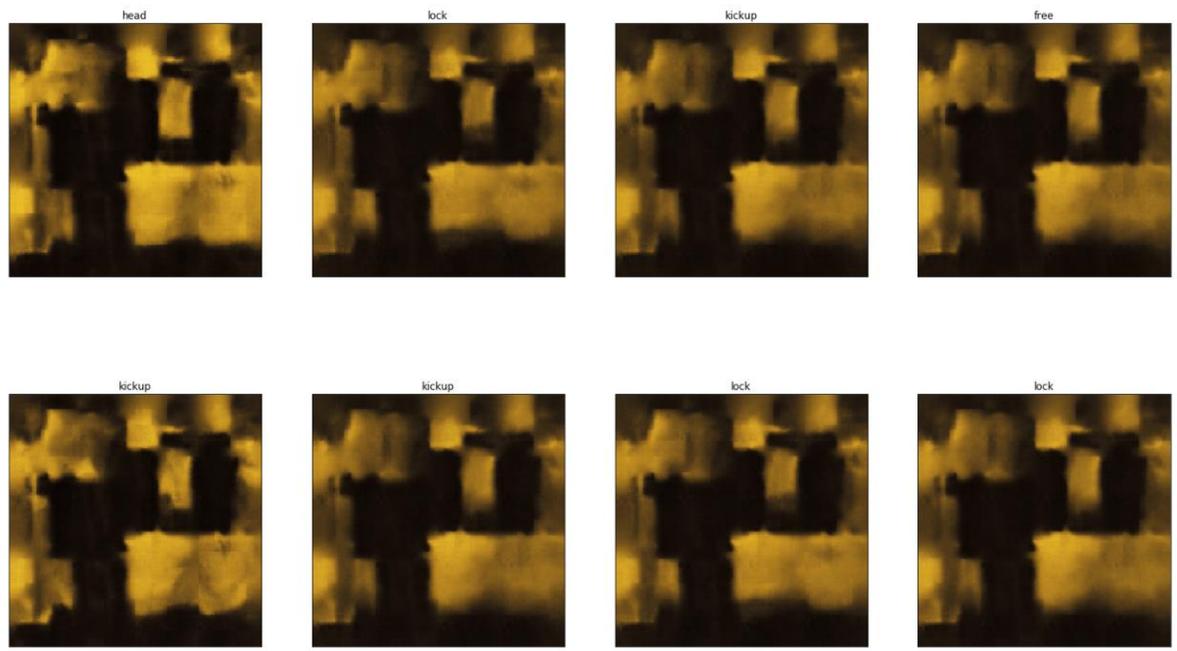

**Fig 37**, Random Latent interpolation of texture images during inference on the CVSSP JP dataset

## 7.7 Conclusion

With this Mesh-texture based CVAE, we have produced a compact representation of both shape and appearance of the 4D video sequence. We have also encoded the Mesh and texture sequence from (1024 x 1024 x 3 (Image) + 4000 x 3 (Mesh) = 3157728) 3157728 vector size to 256 latent vector representation, which is roughly $8 \times 10^{-3}$ reduction in size. The output generated is a fairly decent reconstruction for a small training data (around 200 example of training data from each class label in CVSSP JP dataset) . At g inference, the model was able to render the Mesh sequence at a speed of 0.00679 per second, which means it is capable of delivering more than 90Hz refresh rate for a VR or AR application.

# 8.0 CHAPTER 5: CONCLUSION AND FUTURE DIRECTION

## 8.1 Conclusion

From rendering the 4D video sequence in HoloLens to compact representation and reconstruction of Mesh and texture sequence . In this project, we have tested the feasibility of both the objectives, Rendering of 4D video sequence in the HoloLens was successfully tested on a small 4D video sequence. Compact representation and reconstruction of shape and appearance of the model was also achieved using different variants of Variational autoencoder and the model generated a decent and faithful reconstruction from the small training dataset. Therefore, there is more opportunity to improve the model's performance with more training dataset and domain expertise in the 3D mesh processing.

## 8.2 Future Direction

There are many scopes to improve the model performance and test new capabilities of the HoloLens device and also an opportunity to rendering the compact representation in the HoloLens device

### 8.2.1 Exploring different Mesh representation:

With more domain expertise and deeper understanding of the Mesh processing, we could develop a much better Neural Network model. The 3D vertex co-ordinates are not translational or rotational invariant consequently, the modern is not able to learn large scale deformation. Therefore, by engineering the feature and different mesh representation like deformation. there is high scope to improve the model performance

### 8.2.2 Employing Graph Neural Network layer on Mesh.

Geometric deep learning is new type of deep learning specific catered for non-Euclidean data structure like Graphs and manifolds (Meshes ) . We can take use these new Graph Layer and explore it potential in developing a better and faster model.

### 8.2.3 Exploring capabilities in the HoloLens 2

Explore new functionalities and API with the New version of the HoloLens, and test the feasibility to render the compact representation of 4d video sequence in the new device. either streaming directly or through Azure cloud services .

## 8.3 References

[1] "What Is Extended Reality Technology? A Simple Explanation For Anyone", Forbes.com, 2019. [Online]. Available: https://www.forbes.com/sites/bernardmarr/2019/08/12/what-is-extended-reality-technology-a-simple-explanation-for-anyone/#15f7f3527249. [Accessed: 26- Aug- 2019].

[2] J. Starck, A. Maki, S. Nobuhara, A. Hilton and T. Matsuyama, "The Multiple-Camera 3-D Production Studio", IEEE Transactions on Circuits and Systems for Video Technology, vol. 19, no. 6, pp. 856-869, 2009. Available: 10.1109/tcsvt.2009.2017406.

[3] C. Budd, P. Huang, M. Klaudiny and A. Hilton, "Global Non-rigid Alignment of Surface Sequences", International Journal of Computer Vision, vol. 102, no. 1-3, pp. 256-270, 2012. Available: 10.1007/s11263-012-0553-4.

[4] " Case study - Capturing and creating content for HoloTour ", Docs.microsoft.com, 2019. [Online]. Available: https://docs.microsoft.com/en-us/windows/mixed-reality/case-study-capturing-and-creating-content-for-holotour. [Accessed: 26- Aug- 2019]

[5] "Holoportation - Microsoft Research", Microsoft Research, 2019. [Online]. Available: https://www.microsoft.com/en-us/research/project/holoportation-3/. [Accessed: 26- Aug- 2019].

[6] "Unity development overview - Mixed Reality", Docs.microsoft.com, 2019. [Online]. Available: https://docs.microsoft.com/en-us/windows/mixed-reality/unity-development-overview. [Accessed: 26- Aug- 2019].

[7] "What is mixed reality? - Mixed Reality", Docs.microsoft.com, 2019. [Online]. Available: https://docs.microsoft.com/en-us/windows/mixed-reality/mixed-reality. [Accessed: 26- Aug- 2019].

[8] " Configure a New Unity Project for Windows Mixed Reality ", Docs.microsoft.com, 2019. [Online]. Available: https://docs.microsoft.com/en-us/windows/mixed-reality/configure-unity-project. [Accessed: 26- Aug- 2019].

[9] A. Ioannidou, E. Chatzilari, S. Nikolopoulos and I. Kompatsiaris, "Deep Learning Advances in Computer Vision with 3D Data", ACM Computing Surveys, vol. 50, no. 2, pp. 1-38, 2017. Available: 10.1145/3042064.

[10]A. Krizhevsky, I. Sutskever and G. Hinton, "ImageNet classification with deep convolutional neural networks", Communications of the ACM, vol. 60, no. 6, pp. 84-90, 2017. Available: 10.1145/3065386.

[11] D. Boscaini, J. Masci, E. Rodolà, and M. Bronstein (2016) Learning shape correspondence with anisotropic convolutional neural networks. In NIPS, pp. 3189–3197.

[12] D. Boscaini, J. Masci, E. Rodolà, M. M. Bronstein, and D. Cremers (2016) Anisotropic diffusion descriptors. Computer Graphics Forum 35 (2), pp. 431–441.

[13] M. M. Bronstein, J. Bruna, Y. LeCun, A. Szlam, and P. Vandergheynst (2017) Geometric deep learning: going beyond euclidean data. IEEE Signal Processing Magazine 34 (4), pp. 18–42.

[14] J. Masci, D. Boscaini, M. Bronstein, and P. Vandergheynst (2015) Geodesic convolutional neural networks on riemannian manifolds. In ICCV workshops, pp. 37–45.

[15] A. Sinha, J. Bai, and K. Ramani (2016) Deep learning 3d shape surfaces using geometry images. In ECCV, pp. 223–240.

[16] P. Wang, Y. Liu, Y. Guo, C. Sun, and X. Tong (2017) O-CNN: Octree-based Convolutional Neural Networks for 3D Shape Analysis. ACM Transactions on Graphics (SIGGRAPH) 36 (4).

[17] P. Wang, Y. Liu, Y. Guo, C. Sun, and X. Tong (2018) Adaptive O-CNN: A Patch-based Deep Representation of 3D Shapes. ACM Transactions on Graphics (SIGGRAPH Asia) 37 (6).

[18] Q. Tan, L. Gao, Y. Lai, J. Yang, and S. Xia (2018) Mesh-based autoencoders for localized deformation component analysis. In AAAI.

[19] L. Gao, J. Yang, Y. Qiao, Y. Lai, P. Rosin, W. Xu, and S. Xia (2018) Automatic unpaired shape deformation transfer. ACM Transactions on Graphics 37 (6), pp. 1–15.

[20] J. Zhu, T. Park, P. Isola, and A. A. Efros (2017) Unpaired image-to-image translation using cycle-consistent adversarial networks. In ICCV.

[21] M. Defferrard, X. Bresson, and P. Vandergheynst (2016) Convolutional neural networks on graphs with fast localized spectral filtering. In Advances in NIPS, pp. 3844–3852.

[22] Q. Tan, L. Gao, Y. Lai, and S. Xia (2018-06) Variational autoencoders for deforming 3d mesh models. In CVPR.

[23] A. Ranjan, T. Bolkart, S. Sanyal, and M. J. Black (2018) Generating 3D faces using convolutional mesh autoencoders. In European Conference on Computer Vision (ECCV), pp. 725–741.

[24] D. Dheeru and G. Casey, "UCI Machine Learning Repository", Archive.ics.uci.edu, 2017. [Online]. Available: https://archive.ics.uci.edu/ml/index.php. [Accessed: 26- Aug- 2019].

[25] P. Domingos, "A few useful things to know about machine learning", Communications of the ACM, vol. 55, no. 10, p. 78, 2012. Available: 10.1145/2347736.2347755 [Accessed 26 August 2019].

[26] "CS231n Convolutional Neural Networks for Visual Recognition", Cs231n.github.io, 2019. [Online]. Available: http://cs231n.github.io/. [Accessed: 26- Aug- 2019].

[27] S. Ioffe and C. Szegedy, "Batch normalization: accelerating deep network training by reducing internal covariate shift", International Conference on International Conference on Machine Learning, vol. 37, pp. 448-456, 2015.

[28] T. Salimans and D. Kingma, "Weight Normalization: A Simple Reparameterization to Accelerate Training of Deep Neural Networks", arXiv.org, 2019. [Online]. Available: https://arxiv.org/abs/1602.07868. [Accessed: 26- Aug- 2019].

[29] I. Jolliffe and J. Cadima, "Principal component analysis: a review and recent developments", Philosophical Transactions of the Royal Society A: Mathematical, Physical and Engineering Sciences, vol. 374, no. 2065, p. 20150202, 2016. Available: 10.1098/rsta.2015.0202.

[30] L. Maaten and G. Hinton, "Visualizing Data using t-SNE", Machine Learning Research, vol. 9, pp. 2579-2605, 2008..

[31] M. Tschannen, O. Bachem and M. Lucic, "Recent advances in autoencoder-based representation learning", arXiv:1812.05069, 2018. [Accessed 26 August 2019].

[32] S. Lombardi, J. Saragih, T. Simon and Y. Sheikh, "Deep appearance models for face rendering", ACM Transactions on Graphics, vol. 37, no. 4, pp. 1-13, 2018. Available: 10.1145/3197517.3201401.

[33] Y. Yuan, Y. Lai, J. Yang, H. Fu and L. Gao, "Mesh Variational Autoencoders with Edge Contraction Pooling", arXiv.org, 2019.

[34] Y. Yang, Y. Yu, Y. Zhou, S. Du, J. Davis and R. Yang, "Semantic Parametric Reshaping of Human Body Models", 2014 2nd International Conference on 3D Vision, 2014. Available: 10.1109/3dv.2014.47 [Accessed 26 August 2019].

[35] A. Gilbert, M. Volino, J. Collomosse and A. Hilton, "Volumetric performance capture from minimal camera viewpoints", European Conference on Computer Vision, 2018.

[36] "4d-repository Home", 4drepository.inrialpes.fr, 2019. [Online]. Available: http://4drepository.inrialpes.fr/.

[37] K. Hornik, "Approximation capabilities of multilayer feedforward networks", Neural Networks, vol. 4, no. 2, pp. 251-257, 1991. Available: 10.1016/0893-6080(91)90009-t [Accessed 27 August 2019].